\documentclass[sn-basic,Numbered]{sn-jnl}

\usepackage[nolist, nohyperlinks]{acronym}
\usepackage{booktabs}
\usepackage{graphicx}%
\usepackage{multirow}%
\usepackage{amsmath,amssymb,amsfonts}%
\usepackage{amsthm}%
\usepackage{mathrsfs}%
\usepackage{makecell}
\usepackage[title]{appendix}%
\usepackage{xcolor}%
\usepackage{textcomp}%
\usepackage{manyfoot}%
\usepackage{booktabs}%
\usepackage{algorithm}%
\usepackage{algorithmicx}%
\usepackage{algpseudocode}%
\usepackage{listings}%
\usepackage{todonotes}
\usepackage{wrapfig}

\theoremstyle{thmstyleone}%
\theoremstyle{thmstyletwo}%

\theoremstyle{thmstylethree}%

\begin{acronym}
\acro{AI}{Artificial Intelligence}
\acro{AutoML}{Automated Machine Learning}
\acro{CNN}{Convolutional Neural Network}
\acro{DAG}{Directed Acyclic Graph}
\acro{DARTS}{Differentiable Architecture Search}
\acro{ENAS}{Efficient Neural Architecture Search}
\acro{LGA}{Label-Gradient Alignment}
\acro{LSTM}{Long Short-Term Memory}
\acro{MCTS}{Monte-Carlo Tree Search}
\acro{MDP}{Markov Decision Process}
\acro{ML}{Machine Learning}
\acro{NAS}{Neural Architecture Search}
\acro{NLP}{Natural Language Processing}
\acro{PEB}{Partial Episode Bootstrapping}
\acro{RL}{Reinforcement Learning}
\acro{RWA}{Random Walk Autocorrelation}
\acro{ViT}{Vision Transformer}

\acrodefplural{CNN}[CNNs]{Convolutional Neural Networks}
\acrodefplural{GNN}[GNNs]{Graph Neural Networks}
\acrodefplural{MDP}[MDPs]{Markov Decision Processes}
\end{acronym}

\newcommand{\gciiacknowledgement}[0]{This research received funding from the Flemish Government (AI Research Program).}

\newcommand{\acknowledgement}[0]{\gciiacknowledgement\\\fwoacknowledgement}

\begin{document}

\title[Scalable Reinforcement Learning-based Neural Architecture Search]{Scalable Reinforcement Learning-based Neural Architecture Search}

\author*[1]{\fnm{Amber} \sur{Cassimon}}\email{amber.cassimon@uantwerpen.be}
\author[1]{\fnm{Siegfried} \sur{Mercelis}}\email{siegfried.mercelis@uantwerpen.be}
\author[1]{\fnm{Kevin} \sur{Mets}}\email{kevin.mets@uantwerpen.be}

\affil*[1]{\orgdiv{IDLab - Faculty of Applied Engineering}, \orgname{University of Antwerp - imec}, \orgaddress{\street{Sint-Pietersvliet 7}, \city{Antwerp}, \postcode{2000}, \state{Antwerp}, \country{Belgium}}}

\abstract{In this publication, we assess the ability of a novel \acl{RL}-based solution to the problem of \acl{NAS}, where a \ac{RL} agent learns to search for good architectures, rather than to return a single optimal architecture. We consider both the NAS-Bench-101 and NAS-Bench-301 settings, and compare against various known strong baselines, such as local search and random search. We conclude that our \acl{RL} agent displays strong scalability with regards to the size of the search space, but limited robustness to hyperparameter changes.}


\maketitle

\section{Introduction}
\label{sec:intro}

    Over the past decade, deep learning has made great strides in several fields, ranging from computer vision to natural language processing and \ac{RL} \cite{Krizhevsky2012} \cite{Dosovitskiy2021}. Much of this success has been driven by the search for better neural network architectures \cite{He2016} \cite{Szegedy2015}, which has in turn significantly increased the complexity of state-of-the-art neural network architectures. This trend has led to the creation of automated methods for finding optimal neural network architectures.

    This set of methods is usually referred to as \ac{NAS}. \ac{NAS} has been done using a wide variety of techniques, from evolutionary algorithms \cite{Elsken2018Efficient} to reinforcement learning \cite{Pham2018} and continuous relaxation \cite{Liu2018}.

    Most \ac{NAS} methods used today are single-use methods, where an algorithm is ran once for multiple hours or days, and yields a single architecture. If any of the parameters of the search are changed, such as the search space, the target application,\ldots the search must be repeated, which can be computationally costly. Typically, computational costs also scale with the search space, with larger search spaces requiring significantly more computational resources In this publication, we take the first step towards building a \ac{NAS} system that scales efficiently with the size of the search space, to limit the necessary computational resources.

    We aim to achieve this by learning a searching behaviour, rather than trying to find an optimal architecture for any given problem. In this paper, we focus on the problem setting and the design of the agent. We examine the agent in the NAS-Bench-101 and NAS-Bench-301 settings.

    More precisely, we make the following contributions:

    \begin{enumerate}
        \item We propose a novel \acl{RL}-based \ac{NAS} methodology based on the incremental improvement of neural network architectures
        \item We investigate the effectiveness of our \ac{NAS} methodology on two established benchmarks: NAS-Bench-101 and NAS-Bench-301
        \item We compare against several known strong baseline algorithms including random search and local search, as well as several state-of-the-art algorithms.
    \end{enumerate}

\section{Related Work}
		Over the years, several types of algorithms have been used to attempt to tackle the \ac{NAS} problem. One popular type of \ac{NAS} algorithm are evolutionary algorithms. They have been used by Real et al. \cite{Real2017Large}, and Elsken et al. \cite{Elsken2018Efficient} to succesfully find architectures that can match or exceed the performance of state-of-the-art hand-designed neural network architectures.
		A more recent example of the evolutionary approach is the work of Hendrickx et al. \cite{Hendrickx2022}. They use a modified version of NSGA-Net \cite{Lu2019NSGA} to optimize convolutional neural networks to with the aim of steering a nanodrone towards a flower so the flower can be pollinated. They modified NSGA-Net to include depthwise separable operations, and started their searches from known, well-performing architectures, such as MobileNetV2. Using this approach, they showed that starting from known well-performing architectures can increase search performance, leading to \ac{NAS} algorithms finding good solutions quicker.

		Another approach that has proved popular is continuous relaxation. This was first introduced in DARTS \cite{Liu2018}. DARTS transforms the discrete set of operations that can be assigned to the edges of a computational graph into a continuous one. This allows them to reframe the problem of finding the optimal operation to assign to each edge as a bi-level optimization problem, where the first level constitutes the trainable parameters of the underlying neural network, and the second level constitutes the architectural parameters that parameterize the choice of operation for each edge. Using gradient descent, they are able to find strong performing architectures, with a reasonable computational budget. DARTS has seen many adaptation in recent years, including BHE-DARTS \cite{Cai2023BHEDARTS}, Fair DARTS \cite{Chu2020Fair}, \ldots
		
		\ac{MCTS} is another approach that has gained popularity in recent years \cite{Wang2020Neural} \cite{Zhang2023Semantic}. Usually, in these approaches, an architecture is broken down into a sequence of decisions. One decision is made at every level of the search tree, until a complete architecture has been sampled. A very recent example is GCNAS \cite{Zhang2023Semantic}, which uses \ac{MCTS} to find a strong architecture for the semantic segmentation of multispectral LiDAR point clouds.

    \subsection{\acl{RL}-Based NAS}
		Next, we examine several \ac{RL}-based \ac{NAS} algorithms in more detail.    
        In their paper on MetaQNN \cite{Baker2017Designing}, Baker et al. showcase a tabular Q-learning-based methodology for iteratively designing \acp{CNN} in a chain-structured, macro search space. They consider three different computer vision target domains: CIFAR-10, SVHN and MNIST. This approach models the design process for a neural network as a sequential decision making problem. A complete neural network is considered to consist of a sequence of layers. MetaQNN decides the type of the layer, and some of its parameters. Using this approach, they were able to achieve comparable performance to some state-of-the-art networks of the time. The experiments in this publication were reported to take 8-10 days using 10 GPUs, resulting in a total computational cost of 80-100 GPU-days.

        Zoph et al. published their paper on performing \acl{NAS} with \acl{RL} in 2017 \cite{Zoph2017}. Contrary to MetaQNN, Zoph et al. used a \ac{LSTM}-based \acl{RL} controller to build neural networks. Their agent was capable of operating both in a chain-structured macro search space for designing \acp{CNN} for CIFAR-10, and in a cell-based search space for designing recurrent cells for use on Penn Treebank. Compared to Baker et al.'s incremental approach, Zoph et al. designed entire neural networks in a single timestep, and used their validation accuracy as a reward signal. Using their algorithm, Zoph et al. were able to achieve state-of-the-art performance or close to it for both CIFAR-10 and Penn Treebank. One of the downsides of their methodology was its large computational cost of 800 GPUs over 28 days \cite{Zoph2018Learning}. Especially compared to the 80-100 GPU-days of Baker et al. \cite{Baker2017Designing}, this was a large hurdle for any researcher without access to large computational clusters.

        From this sprang efforts from various researchers, among who were Pham et al., to lower the amount of necessary computational resources to conduct \ac{NAS} research. In their paper ``Efficient Neural Architecture Search via Parameter Sharing" \cite{Pham2018}, they proposed a method to speed up the search process. Similar to Zoph et al., Pham et al. considered the CIFAR-10 and Penn Treebank problems. Pham et al. were able to achieve a speed up by a factor of 1000x compared to Zoph et al.'s first publication \cite{Zoph2017}, resulting in an overall computational cost of between 8 (CIFAR-10, Macro Search Space) and 10 GPU-hours (Penn Treebank  and CIFAR-10 Micro Search Space). Pham et al. were able to achieve this speed-up through the use of parameter sharing, or weight sharing. Realizing that the major bottleneck in \cite{Zoph2017} was the training of neural architectures from scratch to convergence, Pham et al. attempted, through the use of weight-sharing, to accelerate this training process, and thus limit the amount of time necessary to sufficiently train each sampled architecture.

        Despite being originally published in 2018, \ac{LSTM}-based controllers are still often used in applications of \ac{NAS}. One example of this is Li et al.'s work in \cite{Li2023Meta}, where an \ac{LSTM}-based controller is used for designing \acp{GNN} capable of transferring between different graph-based tasks. They do not use weight-sharing, but rather use a performance prediction model to even further increase computational efficiency, and avoid some pitfalls of weight sharing specific to \acp{GNN} \cite{Zhou2022Auto}.

        A more recent method for performing \ac{NAS} using \ac{RL} is GraphPNAS \cite{Li2022Graphpnas}. In this work, a probabilistic graph generator is trained using the REINFORCE algorithm \cite{Williams1992Simple}. The authors show strong performance on various benchmarks, with a computational cost of 16 GPU-days for the Tiny-ImageNet with Oracle Evaluator setting and 12 GPU-hours in the ENAS Macro search space setting on CIFAR-10.

\section{Methods}
	Originally defined in \cite{Elsken_2019_Neural}, we consider a solution to a \ac{NAS} problem to consist of three components: a search space, a search strategy and a performance estimation strategy. In this section, we will consider these three aspects for our method. For a more thorough overview of the different types of search spaces, search strategies, performance estimation strategies etc., we refer interested readers to \cite{white2023neural}.
	
	Besides these aspects, we would also like to briefly make a note of the algorithm we use to identify isomorphic graphs. We adopt a modified version of the graph hashing algorithm used in \cite{Ying2019} to detect isomorphic graphs. This algorithm is used for quick look-ups into the NAS-Bench-101 dataset, it is used to check the uniqueness of architectures when generating neighbours, it is used in our local search and random search algorithms to identify architectures, etc. Concretely, we improved the performance of the algorithm by removing many of the string $\leftrightarrow$ bytes conversions. We also replaced the MD5 hashing algorithm in the original with the blake2s hashing algorithm with a 32-byte digest, due to its increased speed. We verified that the new hashing algorithm doesn't cause any collisions, by computing the hash of every architecture in the NAS-Bench-101 benchmark to verify that our changes do not create any hash collisions. Finally, we also verified our algorithm produces the same outcomes on the unit tests that were a part of the original NAS-Bench-101 codebase. We also note that the new 32-byte digest is twice as long as the 16-byte digest of the original MD5 algorithm, which should in theory reduce the chance of hash collisions.

    \subsection{Search Space}
    \label{ssec:methods:search-space}
        Since \acl{NAS} is a search problem, we will start by defining the search spaces considered in this paper. Specifically, we consider two search spaces, the first is that from NAS-Bench-101 \cite{Ying2019}, and the second is that from NAS-Bench-301 \cite{Siems2021Nasbench}.

        The NAS-Bench-101 search space is a cell-based, operations-on-nodes search space. It consists of 423624 unique directed acyclic graphs, with at most 7 vertices and 9 edges, which represend the computational \ac{DAG} of one cell in a neural network. Each vertex can carry one of five labels: ``input", ``output", ``conv-1x1", ``conv-3x3" and ``max-pool-3x3". Each graph has 1 vertex labelled as ``input" with index 0, and the vertex with the highest index is assigned the "output" label. All vertices in between can be assigned ``conv-1x1", ``conv-3x3" or ``max-pool-3x3". Every graph is required to have at least 1 path going from the vertex labelled ``input" to the vertex labelled ``output", and all vertices must have an in- and out-degree of at least 1 (With exceptions for vertices labelled ``input" or ``output").

		\begin{figure}
            \centering
            \includegraphics[width = \textwidth, clip]{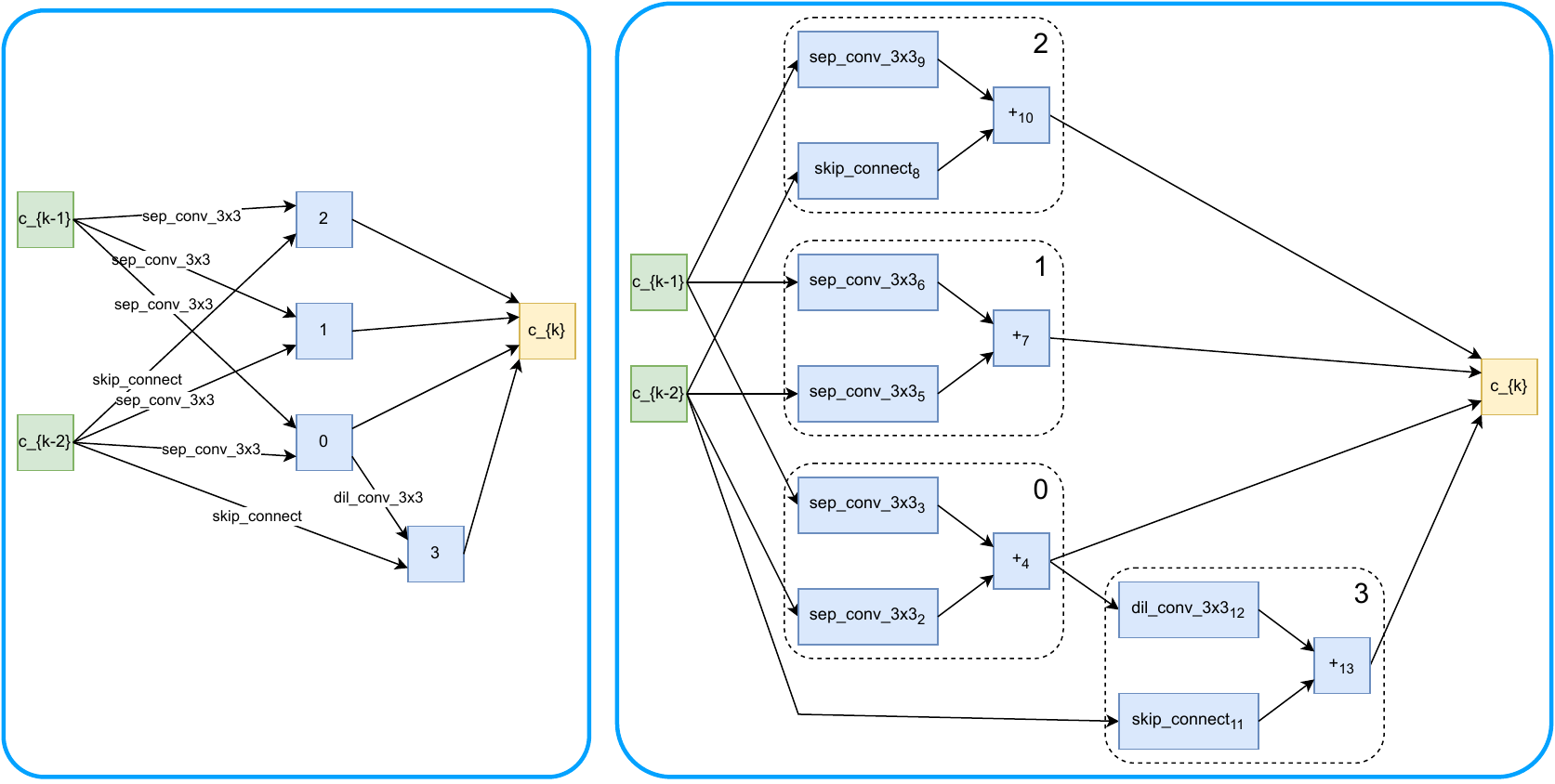}
            \caption{An example of a ``operations-on-edges" architecture (left) converted to a ``operations-on-nodes" representation (right). All edges with associated operations are converted to nodes. Each of these nodes is given an in-edge from the source of the original edge, and an out-edge to the destination of the original edge. The nodes in the original architecture are replaced by reduction operations, a summation in this case. Finally, as was the case before, all reduction operations are given an edge to the output operation.}
            \label{fig:ops-on-edges-nodes}
        \end{figure}

        The NAS-Bench-301 search space is a cell-based, operations-on-edges search space that contains roughly $10^{18}$ architectures. Each cell has 7 vertices, with the first two being labelled as ``input" and the last one being labelled as ``output". The cell also contains 4 intermediate vertices, each with 2 incoming edges. Finally, all 4 intermediate vertices also have 1 edge going to the output vertex, where the feature maps from each intermediate vertex are concatenated along the depth dimension. Each of the 8 incoming edges for the 4 intermediate vertices will have an operation assigned to it, picked from ``avg\_pool\_3x3", ``dil\_conv\_3x3", ``dil\_conv\_5x5", ``max\_pool\_3x3", ``sep\_conv\_3x3", ``sep\_conv\_5x5" or ``skip\_connect". Each of these edges can use as its source any node with an index lower than its destination node (to ensure the computational graph is acyclic.) The two edges incident to one reduction node must also originate from different sources. In total, the described search space contains around $10^{9}$ architectures. This is raised to $10^{18}$ by searching for two architectures from this search space, a normal and a reduction cell. Both of these cells are then used to construct the final neural network. When encoding NAS-Bench-301 architectures for our agents, we convert the operations-on-edges architecture to a operations-on-nodes architecture. An example of this conversion is displayed in figure \ref{fig:ops-on-edges-nodes}.

    \subsection{Search Strategy}
    \label{ssec:methods:search-strategy}
    	The second component of a \ac{NAS} algorithm is the strategy used to traverse the search space. In this section, we will consider this search strategy by first giving a description of the problem our \ac{RL} agent is tasked with solving, along with a brief description of the agent.
    	
    	\subsubsection{Incremental Problem Formulation}
        	We will describe our problem formulation in more detail. For this formulation, we will use the \ac{MDP} framework, which is usually used to describe sequential decision making problems, especially in the field of \ac{RL}. An \ac{MDP} is an 6-tuple $\left(S, A, T, \gamma, \mu, R\right)$ where:

        	\begin{itemize}
            	\item $S$ is the state space
            	\item $A$ is the action space
            	\item $T: S \times A \times S \rightarrow \left[0, 1\right]$ is a probabilistic transition function.
            	\item $\gamma \in \left[0, 1\right)$ is a discount factor
            	\item $\mu: S \rightarrow \left[0, 1\right]$ is a probability distribution over initial states
            	\item $R: S \times A \times S \rightarrow \mathbb{R}$ is a reward function.
        	\end{itemize}

			Sequential decision making problems in \ac{RL} are usually played out in episodes. At the start of an episode, an initial state, $s_{0}$ is selected from the overall state space $S$ using the distribution $\mu$. The agent then observes this state $s_{0}$, and selects an action, $a_{0}$ based on this observation. Once an action is selected, the next state, $s_{1}$ is determined using the transition function $T$, and a reward is computed using $R$. This sequence of observing the current state, selecting an action, and determining the next state and reward is generally referred to as a timestep. The agent plays out one or more timesteps, until either a terminal state is reached, where the episode naturally terminates, or until the episode is truncated for training purposes.

        	In our case, the state space, $S$ is equivalent to the architecture search space, each architecture (or pair of architectures, in the case of NAS-Bench-301) represents one state, $s$ in the \ac{MDP}.
        	In each state $s$, an agent has a set of actions it can take. An action $a$, in our case, corresponds to another architecture or tuple  of architectures in the search space that the agent can move to. The agent is only presented with actions that are reachable by making one change to the current state (For example: Changing the operation of one node into another).
        	The total number of actions that the agent is presented with is limited at a maximum of $N$. If there are less than $N$ neighbours, the invalid actions will be masked-off, and they will not be considered for transition by the \ac{MDP}. The agent is also presented with one additional action that terminates the episode at the current state.
        	Our transition function $T$ is entirely deterministic, the next state is the state the agent indicates through its action.
        	$\gamma$ is a discounting factor that determines what behaviour is considered optimal. Generally, $\gamma$ is somewhere in the $\left[0, 1\right)$ range, with $0.9, 0.95 \textrm{ and } 0.99$ being common values for the parameter. Higher values of $\gamma$ bias policies to more strongly consider future rewards over immediate rewards, while policies found under lower values of $\gamma$ tend to favour immediate and short-term rewards over long-term rewards.
        	$\mu$ determines how the initial state of an episode is selected. In our case, we select an architecture at random from the search space, using the same sampling logic used for our random search algorithm in section \ref{sssec:method:agents:random_search}. 
        	Finally, $R$ is the reward function. In our case, our reward function is defined as the difference between the (validation) accuracy of the previous architecture, and the current architecture. For the first timestep of an episode, since there is no previous architecture, the reward is 0. This difference is computed after applying reward shaping, described in \ref{ssec:reward-shaping}.

        	In essence, we have reframed the \ac{NAS} problem as a graph search problem, where every node in a graph represents an architecture, and every edge represents a relation between architectures. Then, the task our agents must learn to perform, is to find the graph node with the highest accuracy.

        	\begin{figure}
            	\centering
            	\includegraphics[width = \textwidth, clip]{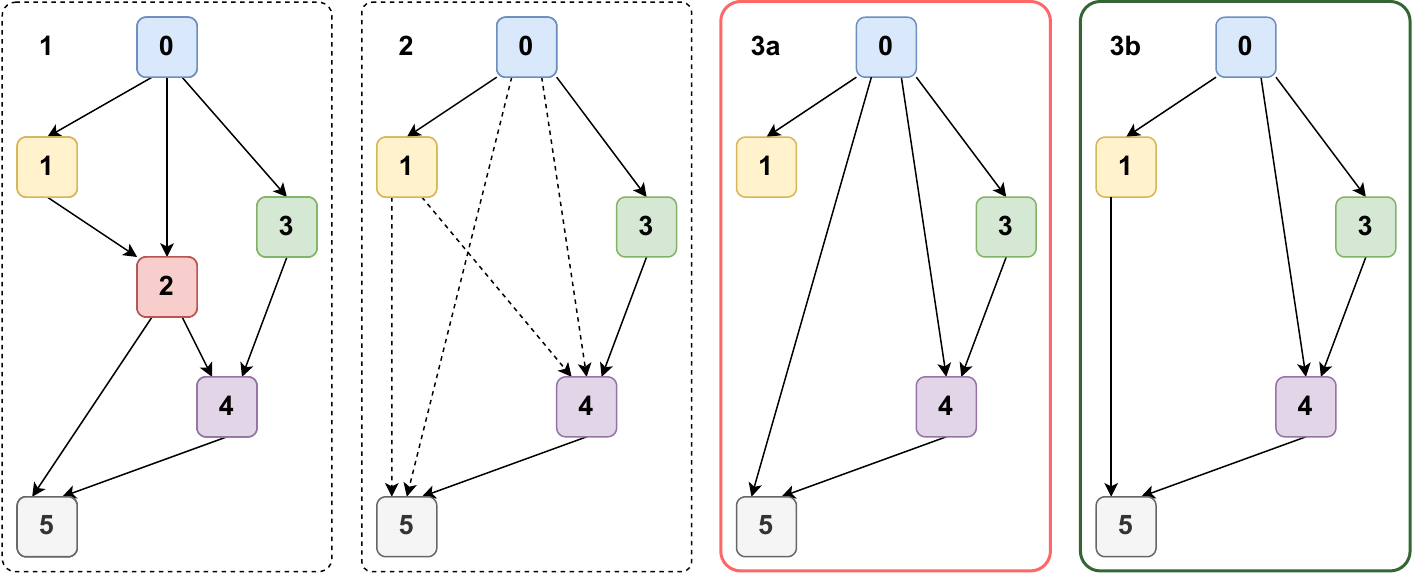}
            	\caption{Vertex Removal Process. 1) A graph consisting of 5 vertices, where we want to remove the vertex with index 2. 2) After removing vertex 2, we generate all possible edges connecting the source of in-edges to vertex 2 to the destination of out-edges to vertex 2. 3a) An invalid selection of generated edges, leaving vertex 1 with an out-degree of 0. 3b) A valid selection of edges, leaving none of the other vertices disconnected from the graph.}
            	\label{fig:vertex-removal}
        	\end{figure}

        	Next, we will describe the process of neighbour generation. Each timestep, an agent is presented with the current architecture, and $N$ of its neighbours, and has to choose one of these architectures as the next state of the \ac{MDP}. If the agent chooses the current architecture, the episode is terminated. These neighbours are generated by making small alterations to the current architecture. Below is a list of the different alterations we use to generate neighbours.

        	\begin{itemize}
            	\item \textbf{Remove a vertex}. When removing a vertex, edges are selected such that the source of in-edges to the removed vertex are connected to the destination of out-edges to the removed vertex. Edges are selected such that all vertices involved have an in- and out-degree of at least 1, to ensure no invalid graphs are generated, this process is illustrated in figure \ref{fig:vertex-removal}. Every possible arrangement of vertices is considered to be a distinct neighbour (after accounting for isomorphism.)
            	\item \textbf{Add a vertex}. The vertex will be connected to one of the preceding and one of the succeeding vertices with an in- and an out-edge respectively. When generating neighbours, all possible ways of connecting the new vertex are generated.
            	\item \textbf{Change a vertex label}. We change the label of a vertex to a different label. This operation can never change a vertex into an input or an output, or change an input or output into a different operation.
            	\item \textbf{Remove an edge}. Only edges that, when removed, don't break the graph's connectivity are considered for removal.
            	\item \textbf{Add an edge}. Edges can only be added between existing vertices, they can not generate cycles, and the total number of edges can never exceed the maximum number of edges of the search space.
        	\end{itemize}

        	In \cite{White2020Local}, White et al. also consider the \ac{NAS} problem through a similar lens. There are some differences though between their framing and ours. In their work, \cite{White2020Local} assume that neighbourhoods are a symmetric relation. That is, if $A$ is in $B$'s neighbourhood, then $B$ must also be in $A$'s neighbourhood. Under our formulation, this isn't necessarily the case. An example of this is shown in figure \ref{fig:vertex-removal}. While the architecture in 3b is considered to be a neighbour of the architecture in 1 (Through removing vertex 2), the architecture in 1 is not a direct neighbour of the architecture in 3b, instead requiring the addition of a vertex, and the addition and removal of several edges. This partly invalidates \cite{White2020Local}'s definition of a branching factor, and thus makes it impossible to apply their theoretical results with regards to the number of local minima etc. to our problem formulation. We also note that our overall search space graph also does not have a regular degree in the way \cite{White2020Local}'s does. A possible solution to resolve these discrepancies would be the inclusion of a ``zeroize" operation in the NAS-Bench-101 search space, akin to the ``zeroize" operation in NAS-Bench-301. This would allow NAS-Bench-101 to be formulated as a \ac{NAS} problem with a fixed topology, where the removal of certain vertices and edges is achieved through the use of ``zeroize" operations, similar to NAS-Bench-301.

        	In some settings, like NAS-Bench-301, more than one cell is designed at a time. In this case, all cells (two in the NAS-Bench-301 case) are shown to the agent at the same time. In order to generate neighbours in this case, we generate the neighbours of each cell individually, and then select up to $N$ tuples, each consisting of 1 neighbour for each cell to form the final selection of neighbours. This leads to a significant increase in the total number of neighbours for benchmarks that search multiple cells at a time. (In the NAS-Bench-301 setting, we noticed that each cell has around 70 neighbours, leading to a total set of around 4900 neighbours, of which 50 are selected for presentation to the agent).

    \subsubsection{Reward Shaping}
		\label{ssec:reward-shaping}
        \begin{figure}
            \centering
            \includegraphics[width = \textwidth, clip]{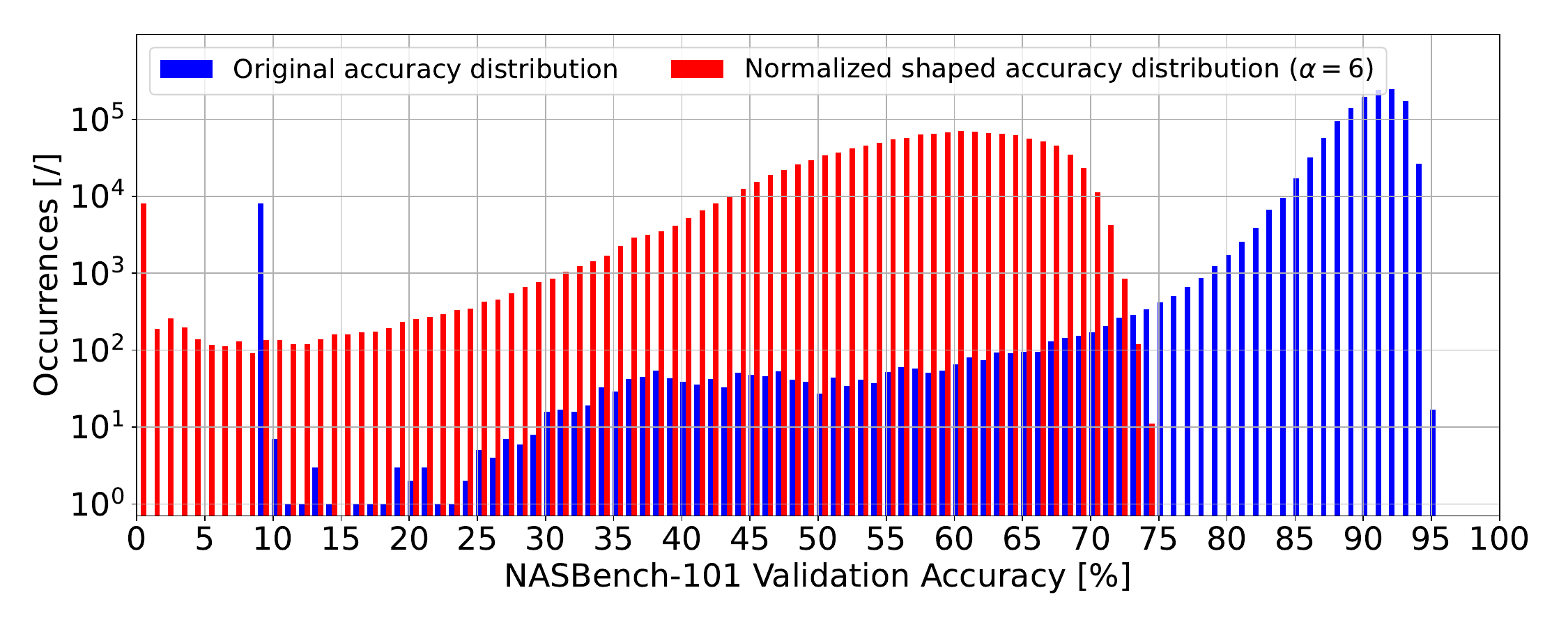}
            \caption{Histogram of the validation accuracy of the architectures included in NAS-Bench-101, across all random initializations. The original accuracy distirbution is shown in blue, while the distribution after reward shaping is shown in red. Note the logarithmic Y-axis.}
            \label{fig:nb101-histogram}
        \end{figure}

        In order for our agent to converge, we utilize reward shaping. In reinforcement learning, reward shaping changes the reward function that is used to train an agent, to facilitate the training process. In our case, we employ reward shaping due to the inherent distribution of rewards in the benchmark problems we consider. In NAS-Bench-101, the average validation accuracy across all random initializations for all architectures is 90.24\% If we plot a histogram of the validation accuracy of all architectures and all random initializations, we also see that the vast majority of architectures have accuracies in the range of $\left[85\%, 94\%\right]$, as demonstrated in figure \ref{fig:nb101-histogram}.
        \begin{wrapfigure}{o}{.5\textwidth}
			\includegraphics[width = .5 \textwidth, clip]{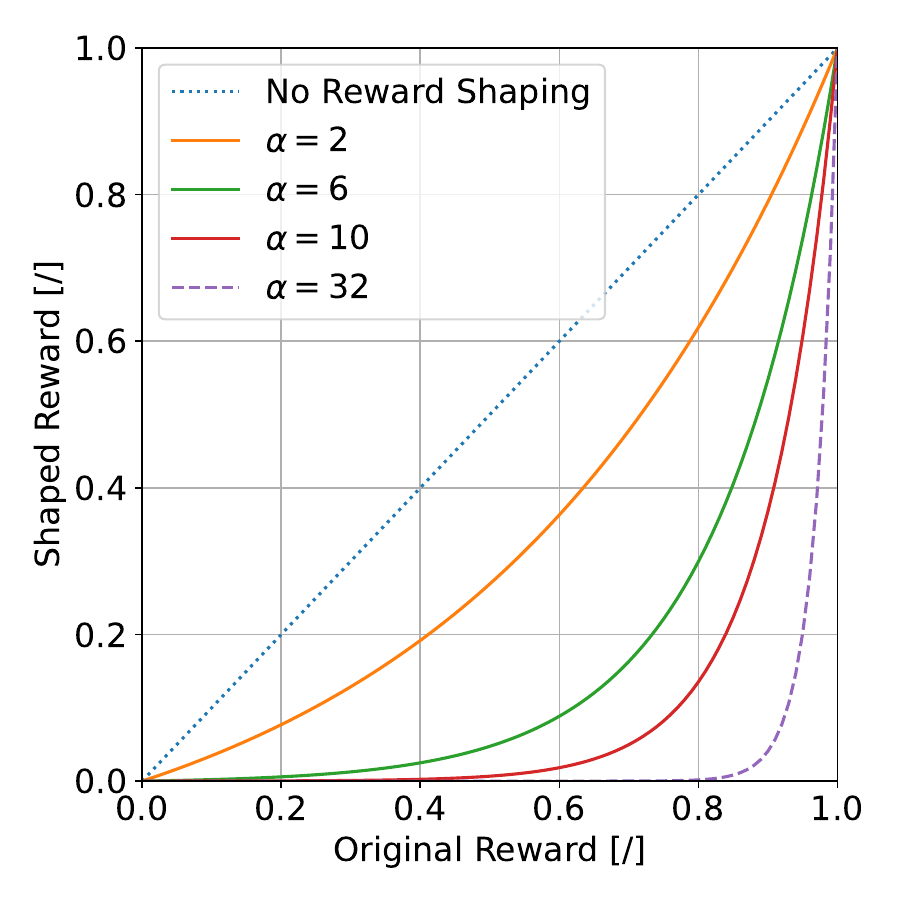}
			\caption{The reward shaping that was used in this paper. Experiments on NAS-Bench-101 used $\alpha=6$, experiments on NAS-Bench-301 used $\alpha=32$. Other values were used in ablation studies.}
			\label{fig:methods:reward-shaping}
	    \end{wrapfigure}
        This means that, even though the reward for our agent is technically normalized between 0 and 1, the majority of architectures will fall into this small range, resulting in very small reward values, after the difference between two consecutive architectures is taken. Thus, we apply an exponential function, $e^{\alpha \cdot R}$, to the validation accuracy of the architectures, and use this exponential accuracy in the difference calculation between subsequent architectures. By selecting an appropriate value for $\alpha$, we can tune the reward function based on the problem definition, to always ensure a good distribution of rewards. Figure \ref{fig:nb101-histogram} shows the difference between the original accuracy distribution, and the distribution after reward shaping. We demonstrate the reward shaping functions used in this publication in figure \ref{fig:methods:reward-shaping}.

    \subsubsection{Agents}
    \label{ssec:methods:agents}
        In order to get a good sense of the performance of our proposed agent, we have also implemented several baseline algorithms. In this section, we will elaborate on the details of each of the agents we evaluate.
    
\begin{figure}
                \centering
                \includegraphics[width = \textwidth, clip]{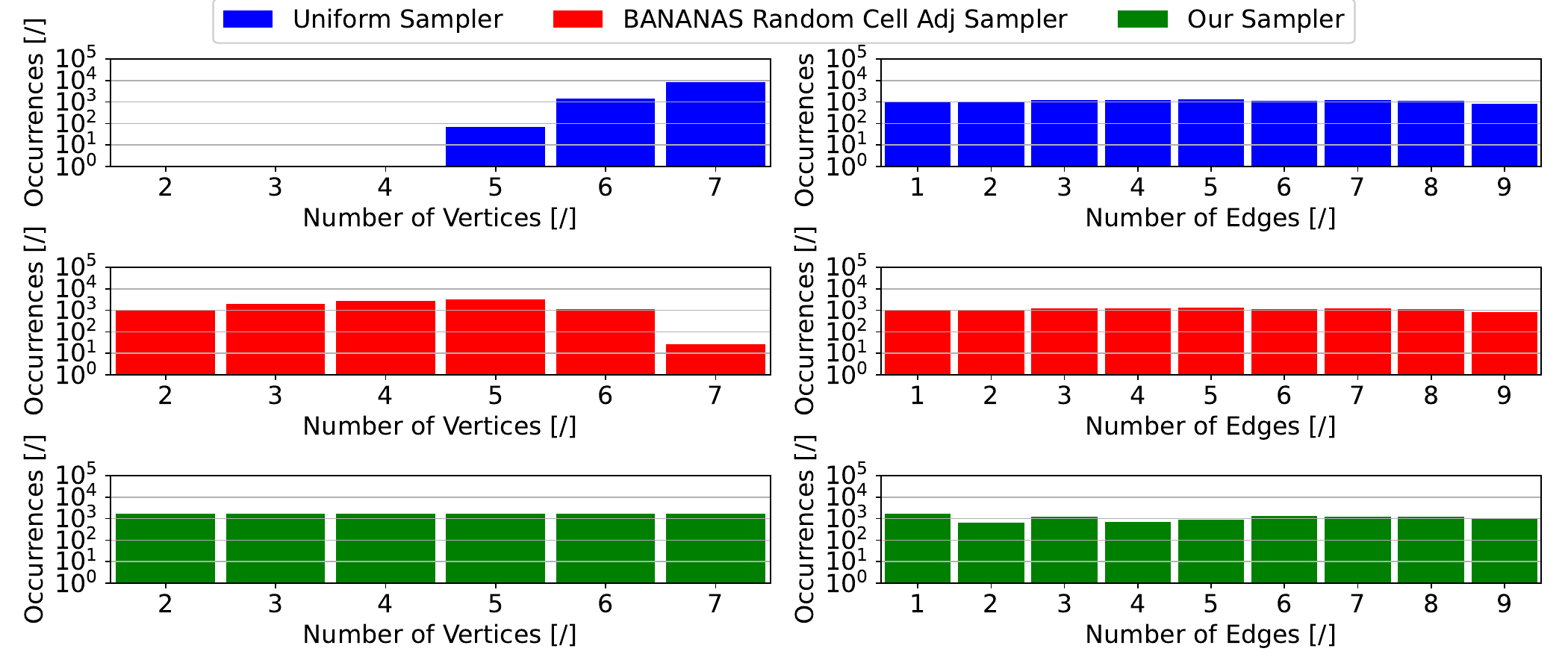}
                \caption{Histogram of the number of vertices and edges in a random sample of 10000 architectures, compared between a uniform sample from the NAS-Bench-101 dataset, the ``Random Cell Adj" sampler from the BANANAS repository, and our sampler. Note the logarithmic Y-axis.}
                \label{fig:random-sampling-histogram}
            \end{figure}

        \paragraph{Random Search}
        \label{sssec:method:agents:random_search}        
            The most basic agent we include is a random search agent. This agent does not follow the incremental problem formulation we intend to use, but rather just selects an architecture at random from the entire search space, similar to the random search agent used in \cite{Li2022Graphpnas}.
            It serves an indication of how much of a change in difficulty or complexity the use of the incremental problem formulation creates. If the incremental problem is significantly more difficult or easy than just picking one architecture from the search space, we expect to see a performance difference between the random search and random walk agent.
            Our random sampling procedure is slightly different from that used in BANANAS \cite{White2021BANANAS}. We start by sampling the size of each graph, in this case from a uniform distribution. Then, we proceed to sample in a manner similar to BANANAS, where we sample random adjacency matrices and vertex labelings, and simply reject the invalid ones until we have a valid sample, with the added requirement that an adjacency matrix must have the desired number of vertices.            
            Figure \ref{fig:random-sampling-histogram} shows a comparison between a uniform sample from the NAS-Bench-101 dataset, the BANANAS random sampler ours is based on, and our sampling algorithm, in terms of the number of vertices of the generated architectures. We note that our sampler produces a quasi-uniform distribution for both the number of vertices and the number of edges, which ensures that our \ac{RL} agent has sufficient experience with both large architectures (many vertices) and smaller architectures (few vertices), thus ensuring consistent performance across the entire search space.

        \paragraph{Random Walk}
            We also include a random agent that follows the incremental problem formulation we defined earlier in section \ref{ssec:methods:search-strategy}. This agent randomly selects a next neighbour according to a uniform distribution. It can additionally also opt to terminate the episode instead of selecting a next neighbour.

        \paragraph{Local Search}
            Some publications \cite{White2020Local}\cite{Ottelander2021} note that local search is a strong \ac{NAS} algorithm, especially in smaller search spaces like NAS-Bench-101. With this in mind, we also include local search as one of the algorithms we benchmark against. Our local search algorithm doesn't have any of the enhancements that \cite{White2020Local} used. The local search agent is presented with $N$ architectures, it compares the validation accuracy for all of them, and selects the one with the highest validation accuracy. If no architecture has a validation accuracy higher than that of the current architecture, the episode is terminated. We note that, different from \cite{White2020Local}, our neighbour relation is non-symmetrical, meaning that A being a neighbour of B, doesn't guarantee that B will also be a neighbour of A, as we explained in section \ref{ssec:methods:search-strategy}.

        \paragraph{\ac{RL}-based Agent}
        \label{sssec:method:agents:rl_agent}
            \begin{figure}
                \centering
                \includegraphics[width = \textwidth, clip]{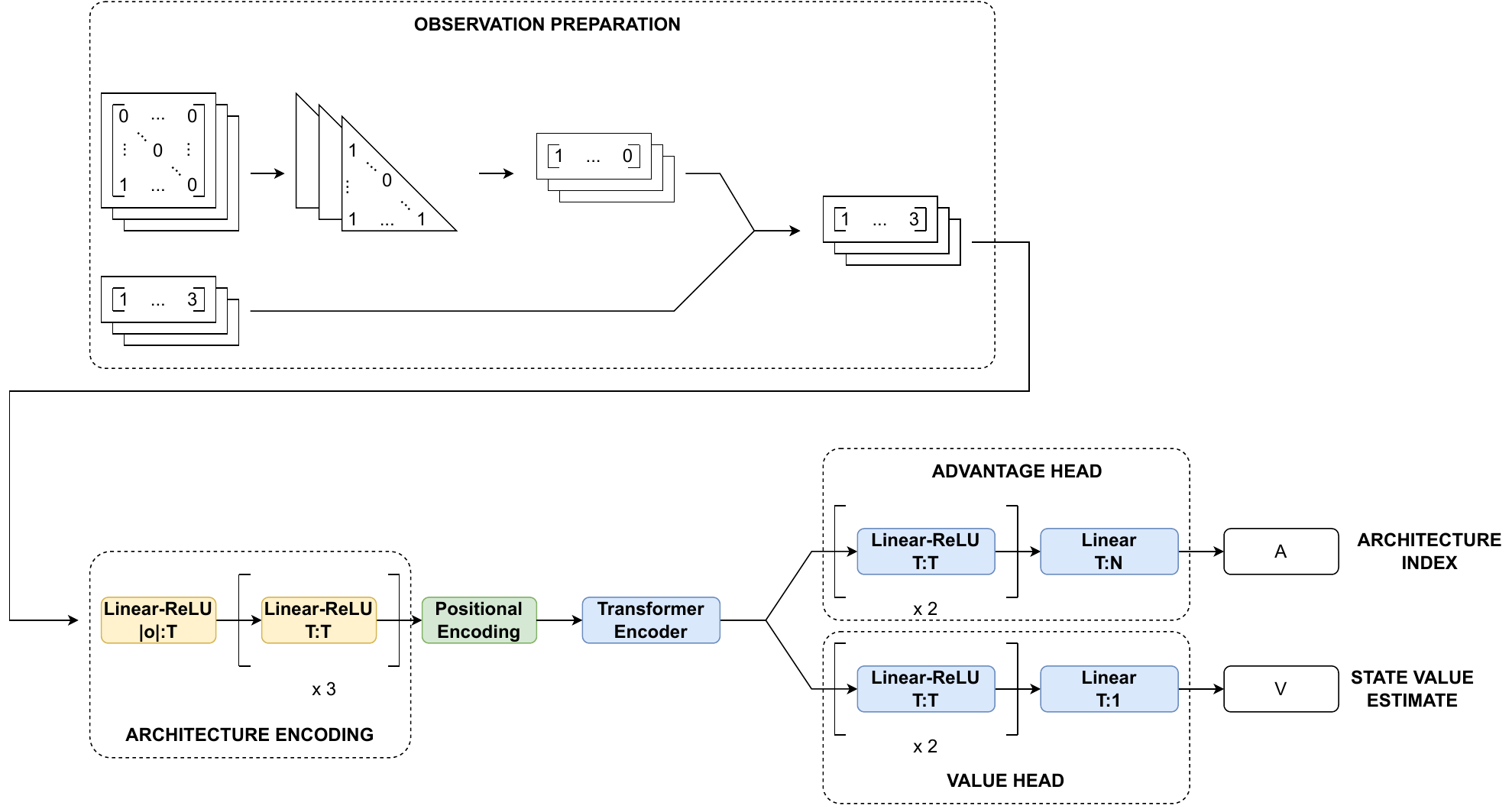}
                \caption{The architecture of our transformer agent. The top row shows the process of preparing the observations, while the bottom row represents the learnable \ac{RL} agent.}
                \label{fig:transformer-arch}
            \end{figure}

            Our \ac{RL} agent, shown in figure \ref{fig:transformer-arch}, uses a transformer encoder as the core of its architecture, similar to \ac{ViT} \cite{Dosovitskiy2021}. The agent is presented with at most $N$ different architectures. Before the architectures are presented to the agent, they must be prepared. We take the lower-triangular half of the adjacency matrix (To ensure acyclicity), and flatten it by concatenating all of the rows. This flattened adjacency matrix is then concatenated to the vertex labels, encoded in a one-hot fashion, resulting in one long binary vector. In order to be able to encode architectures with varying sizes, the original adjacency matrix and vertex labels are first padded to the size of the largest architecture. If multiple architectures need to be searched (As in the case of NAS-Bench-301), the architectures are concatenated after preparation. Before sending the architectures to be encoded, we prepend the the current architecture to the sequence of up to $N$ neighbours.

            After preparation, an architecture encoding network transforms the prepared observation to a 256-dimensional latent space ($T = 256$), using several fully-connected layers with ReLU activations. These latent space vectors then have a positional encoding applied to them, to ensure that the permutation-invariant transformer architecture remains aware of the order of the latent space vectors. While the exact ordering of the architectures is unimportant, the agent must still be aware of their ordering, since each action the agent can perform corresponds to one of the architectures observed. Thus, if the agent loses all awareness of the ordering of all architectures, it will be unable to select the appropriate action.

            The current architecture is prepended to all possible neighbours, similar to the ``classification token" in \cite{Dosovitskiy2021}. Correspondingly, we use the first output of the transformer encoder as the input for the rest of the agent. At this point, we have a duelling head consisting of two branches \cite{Wang2016Dueling}. One uses the transformer output to compute advantage values for each architecture that was presented to the transformer encoder, which is used to determine which action the agent will take. The second branch computes a state-value, and is used to stabilize the \ac{RL} procedure.
            
            The agent is trained using the Ape-X algorithm \cite{Horgan2018Distributed}, a variant of Q-Learning designed for a high throughput of experience collection. We enhance Ape-X using \ac{PEB} \cite{Pardo2018Time} and 3-step bootstrapping to be able to train the agent in finite-length episodes, while obtaining a behaviour suitable for episodes of infinite length. Our Q-learning network makes use of a target network \cite{Mnih2013Playing}, duelling head \cite{Wang2016Dueling} and double Q-learning \cite{Hasselt2010Double}.

	\subsection{Performance Estimation}
		For this publication, we consider two benchmark problems: NAS-Bench-101 and NAS-Bench-301. We selected NAS-Bench-101, since it is the largest tabular benchmark we are currently aware of. Being a tabular benchmark, it eliminates some possible confounding factors that may arise when using other performance estimation strategies. We also included NAS-Bench-301, since it covers a very large and commonly used search space ($\approx 10^{18}$ architectures). Contrary to NAS-Bench-101, NAS-Bench-301 uses a machine learning model fitted on a subset of the search space to predict the performance (classification accuracy, in this case) for the entire search space. All agents we trained use the same performance estimator, in order to elimate the choice of performance estimator as a possible confounding factor.

\section{Experiments}
\label{sec:experiments}
    In order to evaluate the effectiveness of \ac{RL} agent, we will evaluate it on two \ac{NAS} benchmarks: \acs{NAS}-Bench-101 \cite{Ying2019} and NAS-Bench-301 \cite{Siems2021Nasbench}.
    We start by evaluating on \acs{NAS}-Bench-101. In their work, White et al. \cite{White2020Local} note that because the search space for \acs{NAS}-Bench-101 is relatively small, relatively simple algorithms like local search tend to perform fairly well.
    It is with this in mind that we selected \acs{NAS}-Bench-301 as a second benchmark. Since it uses the DARTS \cite{Liu2018} search space, which many other \ac{NAS} algorithms also use, it should give us a clearer view of how our agent performs compared to algorithms like local search.

    \subsection{NAS-Bench-101}
        \subsubsection{Experiment Configuration}
            When tackling NAS-Bench-101, we train our agents for $10 \times 10^{6}$ timesteps of experience. As a reward signal, we use the mean of all three validation accuracies after 108 epochs of training (Corresponding to the reduced noise setting in \cite{White_2021_Exploring}). We train 5 agents with seeds ranging from 0 to 4 (inclusive), using version 1.13 of the RLLib  framework \cite{Liang2018RLLib}. We set $\gamma = 0.9$ with a learning rate of $4 \times 10^{-5}$.
            Our replay buffer has a capacity of $25 \times 10^{3}$ entries. Our replay buffer is a prioritized replay buffer \cite{Schaul2016Prioritized} with $\alpha=0.6$ and $\beta=0.4$.
            Exploration is done using a per-worker epsilon-greedy strategy. We use the same parameters as section 4.1 of the Ape-X paper \cite{Horgan2018Distributed}: Each individual worker uses an epsilon-greedy exploration strategy, with $\epsilon_{i} = \epsilon^{1 + \left(\alpha \cdot \frac{i}{N - 1}\right)}$. We set $\epsilon = 0.4$ and $\alpha = 7$, these are the RLLib default settings.
            We train our agent using episodes of 16 timesteps, and evaluate it with episodes consisting of 32 timesteps. We refer to figure 6 in \cite{Ying2019}, plotting the \ac{RWA} in the NAS-Bench-101 search space. We note that after a 6-10 timestep random walk through the search space, \ac{RWA} flattens out close to zero, indicating that there should be no more locality effects. This implies that, in order for agents to make significant alterations, the maximum episode length should be greater than 6, supporting our choice for 16.
            As mentioned in section \ref{ssec:reward-shaping}, we use reward shaping with $\alpha=6$.
            Our experimental setup allocates 8 threads for 8 workers to gather experience, 4 threads for 4 shards of the replay buffer, and 2 threads for the driver, as well as 1 GPU for the driver, but not for the workers. Different experiments were carried out using either 1 NVIDIA Tesla V100 GPU, or 1 NVIDIA Quadro RTX4000 GPU. Each worker also uses a vectorized version of our environments, that allows it to operate on 32 environments at a time.
            For evaluation, we randomly selected 5 sets of $1 \times 10^{4}$ architectures from the search space. These are used as the initial state for each episode of the evaluation process. Each of the 5 trained agents is evaluated on each of the 5 sets of $1 \times 10^{4}$ initial states, resulting in a total of $2.5 \times 10^{5}$ episodes of evaluation data per algorithm. For evaluation of NAS-Bench-101, we use the image classification accuracy on the test set after 108 epochs of training.

        \subsubsection{Results}
        \label{sssec:nb101-results}
			In the NAS-Bench-101 setting, the mean training time was $92.57$h ($\sigma=20.89$h), with a minimum of $78.27$h, and a maximum of $133.97$h. During evaluation, using only CPUs (no GPUs), completing an episode took an average of $0.50$s ($\sigma=0.49$ s), with a minimum of $0.00$s and a maximum of $10.14$s.        

			\begin{figure}
                \centering
                \includegraphics[width = \textwidth, clip]{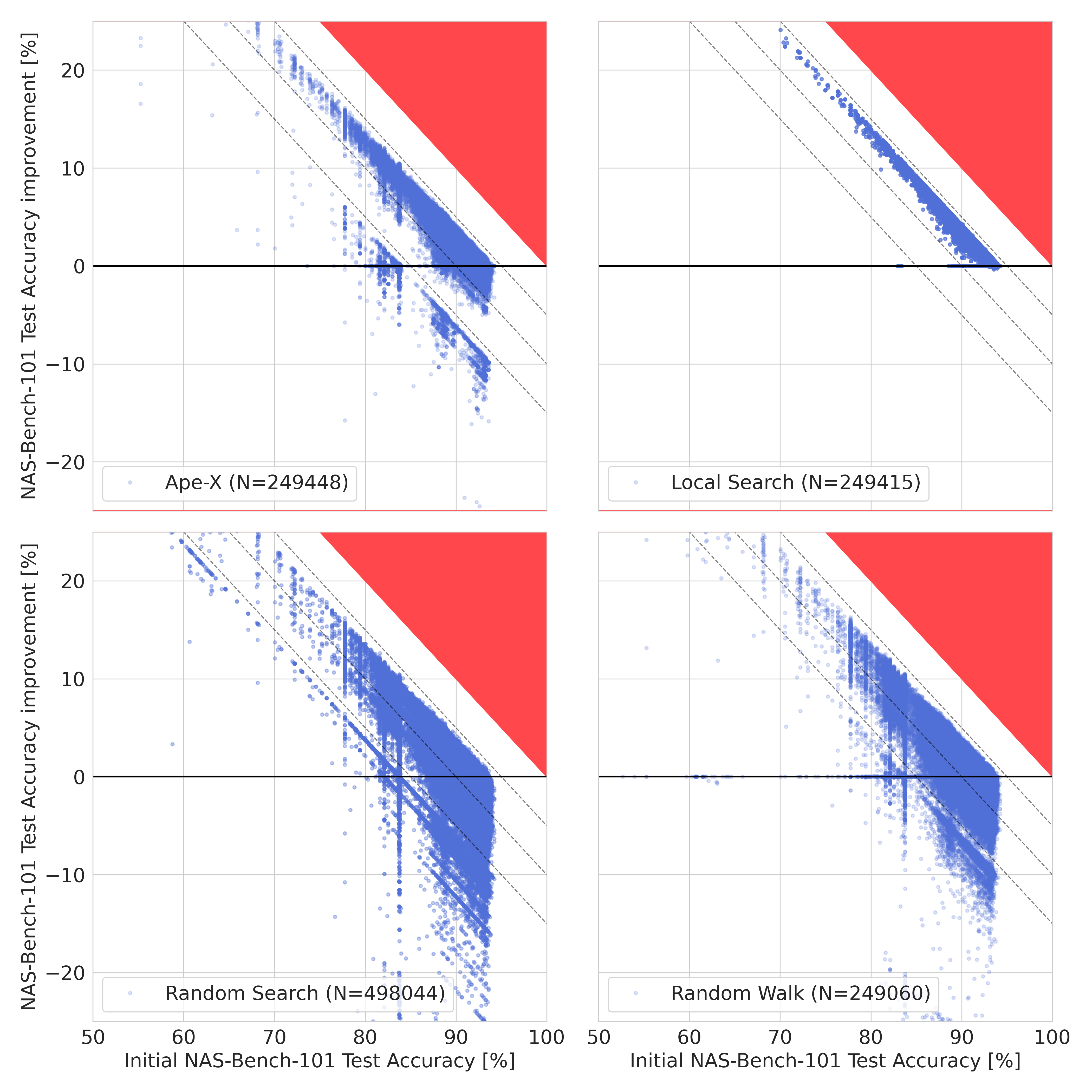}
                \caption{The initial accuracy of the agent vs the improvement in accuracy. The diagonal dashed lines represent a final accuracy (at the end of an episode) of 85\%, 90\% and 95\% respectively. The diagonal line delimiting red region in the top-right corner represents an accuracy of 100\%.}
                \label{fig:nb101-results:initial-vs-improvement}
            \end{figure}        

           We start by evaluating the ability of our agents to improve from a given starting accuracy. We do this by plotting the improvement in accuracy over an entire episode against the accuracy of the initial state. An ideal agent would generate a diagonal line, as close as possible to the top right corner. How close this diagonal can be to the top-right corner is determined by the global optimum within the search space. An agent that consistently manages to end up at the same architecture (Regardless of its performance), will have all datapoints on a perfect diagonal line that intersects the X-axis at the accuracy of this final architecture. Thus, the spread of an agent's datapoints from a perfect diagonal line can be used to gauge how consistently an agent manages to find the same architecture. A deviation from a diagonal line indicates an agent that doesn't always perform consistently, and can optimize some architectures better than others.
           These results are shown in figure \ref{fig:nb101-results:initial-vs-improvement}. From this, we can see that every tested policy displays some degree of inconsistency, as indicated by none of the policies forming a perfect diagonal line. As expected, this is greatest for random policies, while local search is the most consistent. Our reinforcement learning agent is fairly consistent in improving the accuracy of the architectures it is given, but sometimes fails to improve, and makes things worse. This can be an issue when using the agent as part of a hyperparameter tuning pipeline, but the agent improves consistently enough that this should be a relatively rare occurrence, as evidenced by some of the other data in this section.

            \begin{figure}
                \centering
                \includegraphics[width = \textwidth, clip]{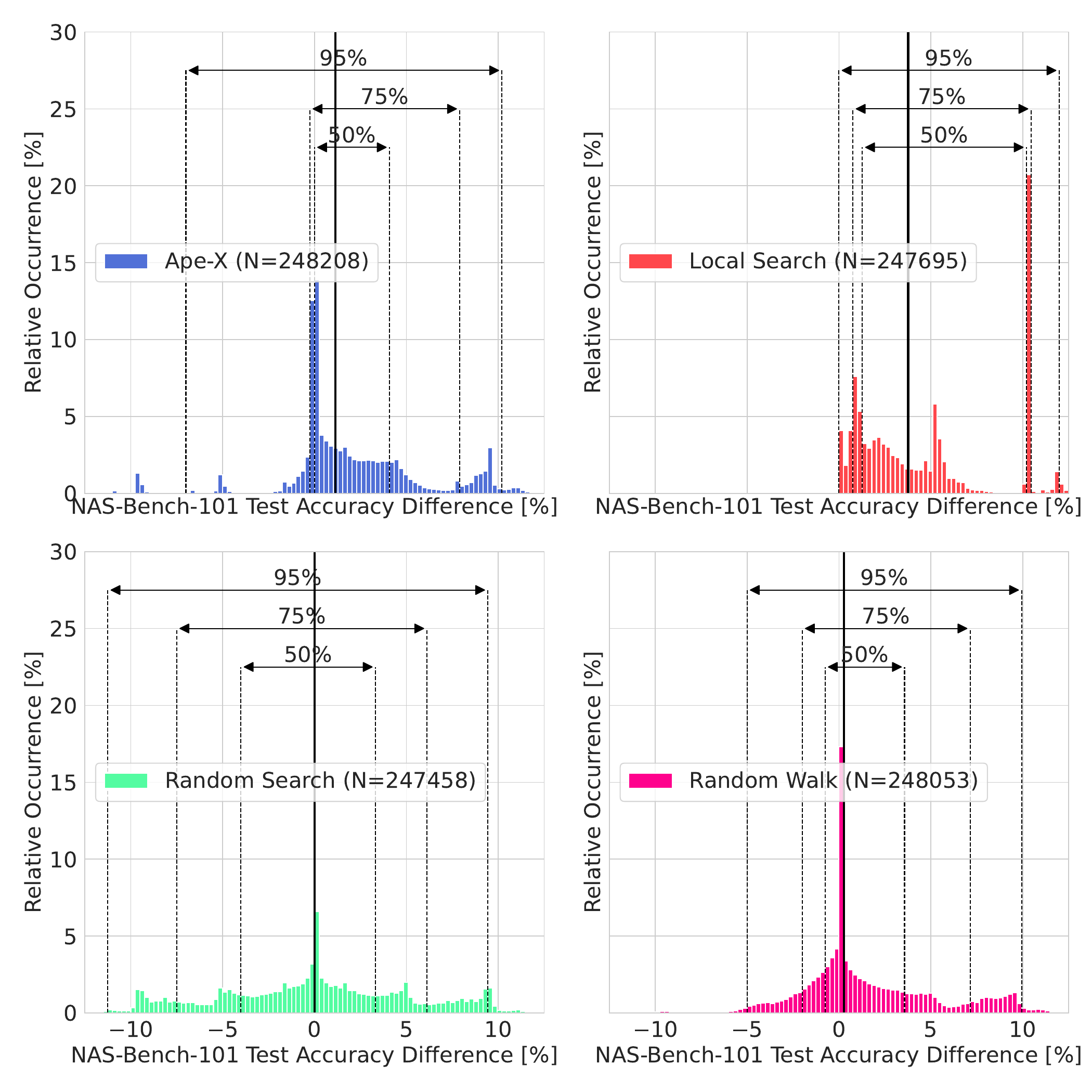}
                \caption{A histogram showing the distribution of the improvement in accuracy over an entire episode. We limited the range of improvements shown between -12.5\% and +12.5\%, which leads to a small amount of outlying samples (Roughly 2000-3000 samples, or 0.8\% - 1.2\% of all samples) being excluded from the histogram (Not included in the extreme bins.) The solid vertical line indicates the median improvement. Dashed vertical lines indicate the edges of a symmetrical range around the median containing a certain percentage of the total samples.}
                \label{fig:nb101-results:improvement-histogram}
            \end{figure}

            We can also look at this data from a different perspective, by evaluating how often each agent is able to improve a given architecture, and how often an agent returns a worse architecture. We perform this evaluation by creating a histogram showing the distribution of the difference in accuracy between the start of an episode, and the end of that episode, where positive values indicate an improvement from the initial to the final architecture. Since the agent decides when to terminate the episode, it should ideally always terminate an episode when the accuracy is higher than the initial state. We show these results in figure \ref{fig:nb101-results:improvement-histogram}.

			\begin{wrapfigure}{O}{.5\textwidth}
            	\centering
            	\includegraphics[width = .5\textwidth, clip]{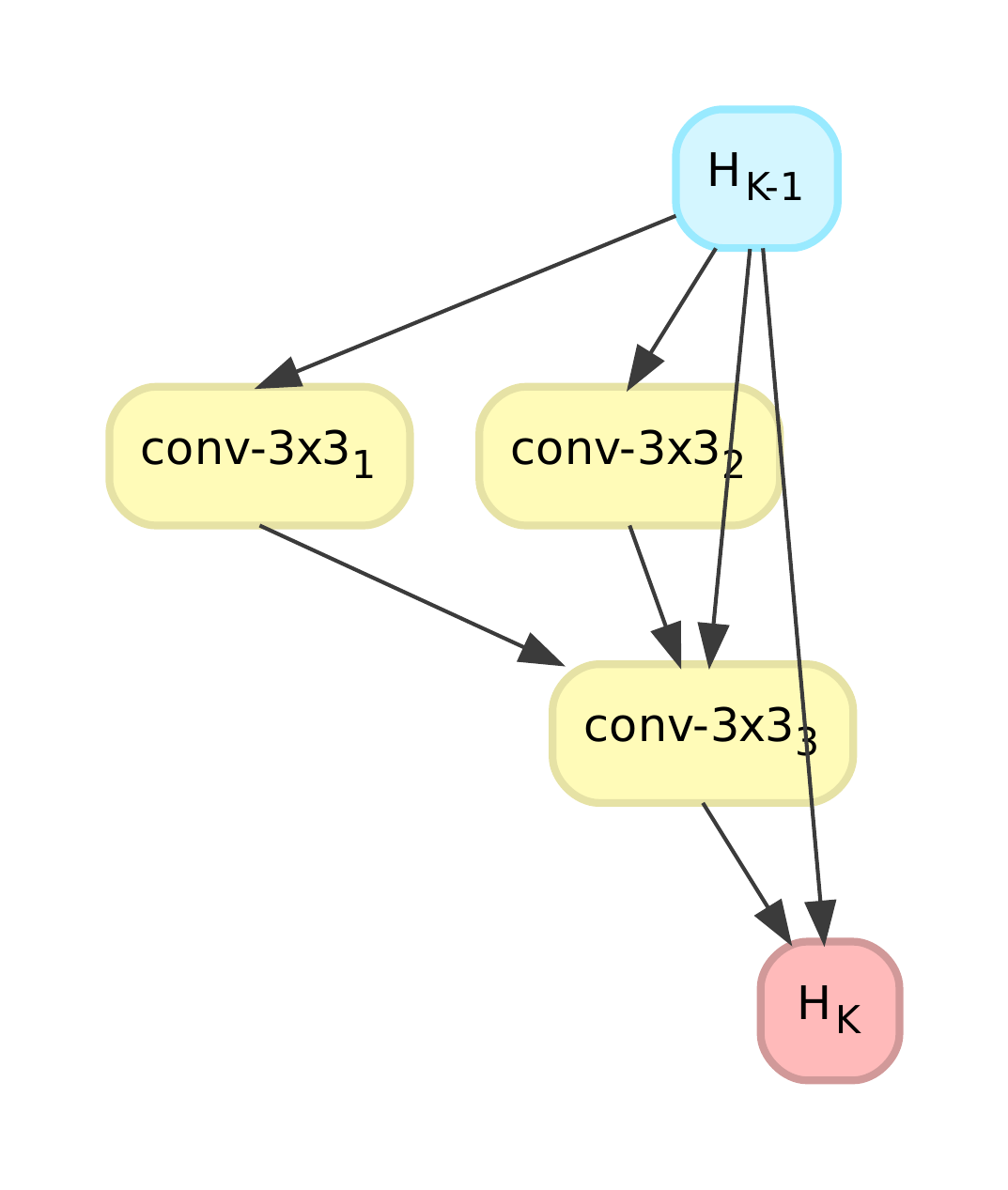}
            	\caption{The local optimum found by local search in the NAS-Bench-101 setting in slightly more than 20\% of cases.}
            	\label{fig:nb101-results:local-optimum}
        	\end{wrapfigure}            
            
            We also show the median value for each histogram using a solid, vertical black line. Besides this, we also show symmetrical intervals around the median. For instance, the interval marked 50\% contains the values between the 25-th and 75-th percentile, i.e., 50\% of values concentrated around the median.
            From this histogram, we can immediately see that, in terms of consistent improvements, local search is undisputedly the best algorithm (Median improvement: $+3.77\%$), followed by Ape-X ($+1.15\%$), Random Walks ($+0.26\%$) and Random Search ($+0.00\%$). This is in line with our expectations, since the local search algorithm is formulated in such a way that it should always improve an architecture. We do note that this isn't always the case, since local search optimizes validation accuracy, but is evaluated on test accuracy. We also note that there are several high peaks in the local search histogram, with one bin even containing around 20\% of the episodes. We found that in this bin, the final architecture is the same in the vast majority of episodes, implying that this architecture is a local optimum. We show this architecture in figure \ref{fig:nb101-results:local-optimum}.
            Quantitatively, we can also see this reflected in the skew value for each distribution. Negative skew values imply that the tail of the distribution is on the left side, while positive skew values indicate the tail is on the right side. Or in other words, negative skew implies that most probability mass is on the right side of the distribution, while positive skew implies the majority of probability mass is on the left side of the distribution. Local search exhibits the lowest skew, with $-2.091$, followed by Ape-X with $-1.821$, Random Walks at $0.091$ and finally random search with a skew of $0.219$.
            It also shows that our \ac{RL} agent most commonly improves an architecture, but on rare occasions fails to do so.
            Finally, one more artifact from this data is the fact that random walks actually outperformed random search, with a median that skews slightly towards the positive side. This becomes even more obvious when we consider the skew by the intervals noted in the histogram. While for random search, most intervals are roughly symmetrical around the median, for random walks, the intervals skew heavily towards the side of positive improvements, with the 95\% interval ranging from -5\% to +10\%.
            Initially, we hypothesized that there may be a correlation between an architecture's test accuracy, and the number of neighbours it has. This data is shown in a scatter plot in figure \ref{fig:nb101-test-accuracy-vs-neighbours}, but unfortunately reveals only a very weak correlation, with Pearson's r at -0.014, and Kendall's $\tau$ at -0.157. We were unable to find another explanation for this performance difference.

            \begin{figure}
                \centering
                \includegraphics[width = \textwidth, clip]{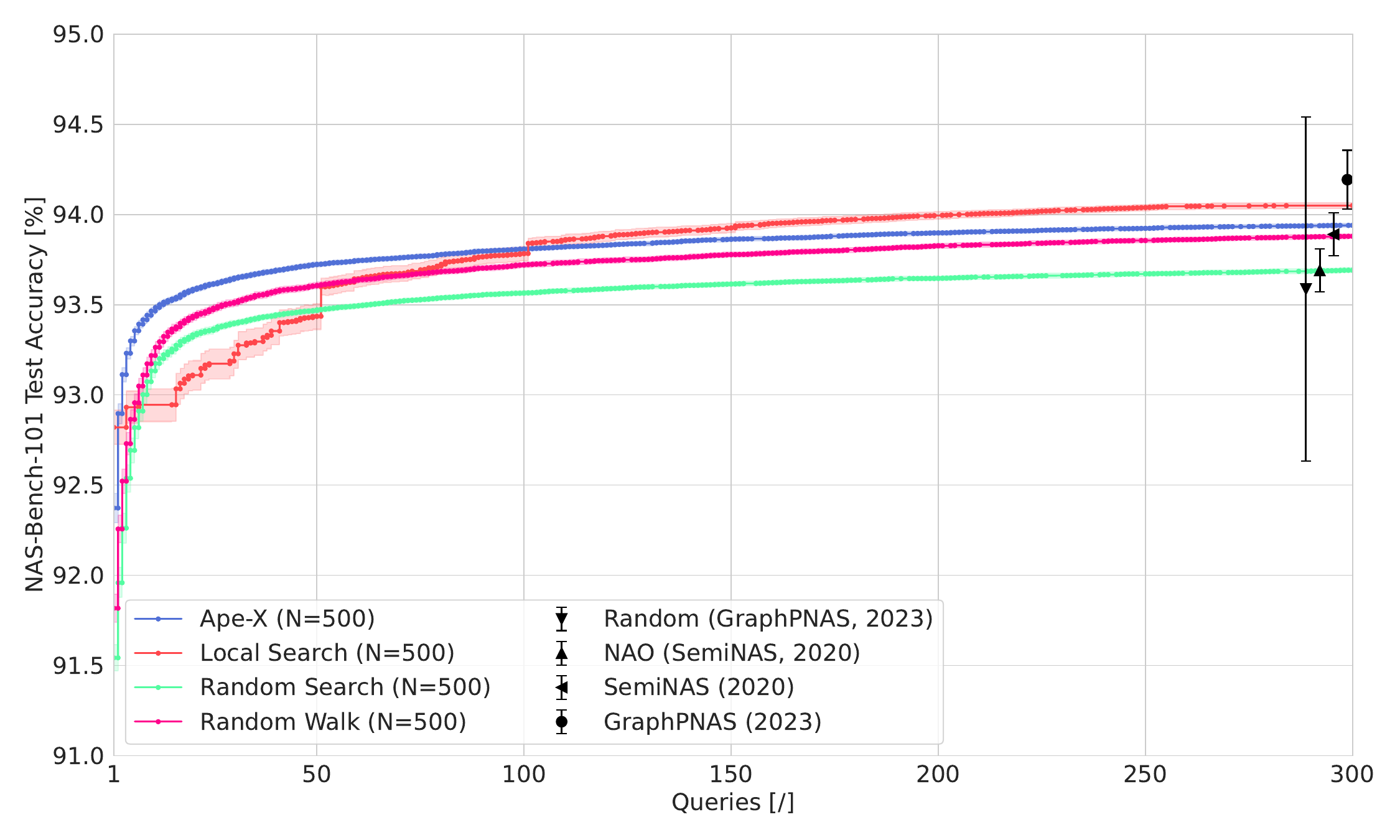}
                \caption{The accuracy of the best architecture found so far plotted against the number of queries. We evaluated each algorithm for up to 300 queries. For baselines indicated  by black lines, we used the numbers reported in their publications. The shaded area (for our experiments) and the error bars (for other baselines) represents a 95\% confidence interval around the mean.}
                \label{fig:nb101-results:best-after-N}
            \end{figure}

            Finally, we also consider the question of how many queries are required for an agent to be able to find well-performing architectures. We consider a setting where an algorithm can make up to 300 queries, and assess the agent by looking at the highest accuracy architecture that has been found after a given number of queries. We note that, for all algorithms except local search, 1 episode is considered to be 1 query. For local search, we count every neighbour at every timestep as a query, since the local search agent needs to know the accuracy of each neighbour, before deciding on which action to take. The results of this are shown in figure \ref{fig:nb101-results:best-after-N}. We also compare against a number of state-of-the-art publications that also reported their results after 300 queries: NASWOT \cite{Mellor2021Neural}, GraphPNAS \cite{Li2022Graphpnas}, Random Search (from GraphPNAS), NAO (Reported by SemiNAS) \cite{Luo2020Semi} and SemiNAS \cite{Luo2020Semi}. For both our own experiments, and the state-of-the-art algorithms, we computed a 95\% confidence interval, to determine if there is a statistically significant difference in performance between algorithms.
            The 95\% confidence interval for our own experiments was computed using bootstrapping with $5 \times 10^{3}$ bootstrap samples, while the confidence intervals for other baselines (Black lines in figure \ref{fig:nb101-results:best-after-N}) were computed using a closed-form expression that assumes a normal distribution, since we don't always have access to the full dataset, but most publications do report a mean and standard deviation. In figure \ref{fig:nb101-results:best-after-N}, we see that our \ac{RL} agent performs fairly well for low query budgets, but as soon as the total number of queries exceeds 75-100, it is outperformed by a local search agent. Local search is known to perform well in the NAS-Bench-101 setting, due to the relatively limited size of the search space \cite{White2020Local}.
            Interestingly, at lower query budgets ($<50$ queries, random walks and random search actually outperform local search. We attribute this to the fact that we allow agents to observe at most 50 neighbours. Thus, if local search is presented with an architecture that has 50 or more neighbours, it first must make 50 queries (one for each neighbour) before it is able to realize an improvement. This also explains the jump in performance that occurs at the 50 query mark for local search.
            Interestingly, we note that, when considering the number of queries, random walks actually outperform random search quite significantly, and even approach the performance of our \ac{RL} agent as the number of queries increases. This may be another indication that not all forms of random search are created equal, and the way a \ac{NAS} problem is formulated can have an impact on how well different random algorithms are able to perform.
            Our first hypothesis for this discrepancy was that larger architectures (More vertices and edges) tend to have both more neighbours and more trainable parameters, and thus a higher accuracy. However, when plotting the number of neighbours an architecture has vs. its test accuracy, it becomes clear that the correlation is rather weak, implying there is likely at least one more factor contributing to random walks strong performance. Unfortunately, we were unable to find any other parameters that may contribute to random walks' strong performance.

            \begin{figure}
                \centering
                \includegraphics[width = \textwidth, clip]{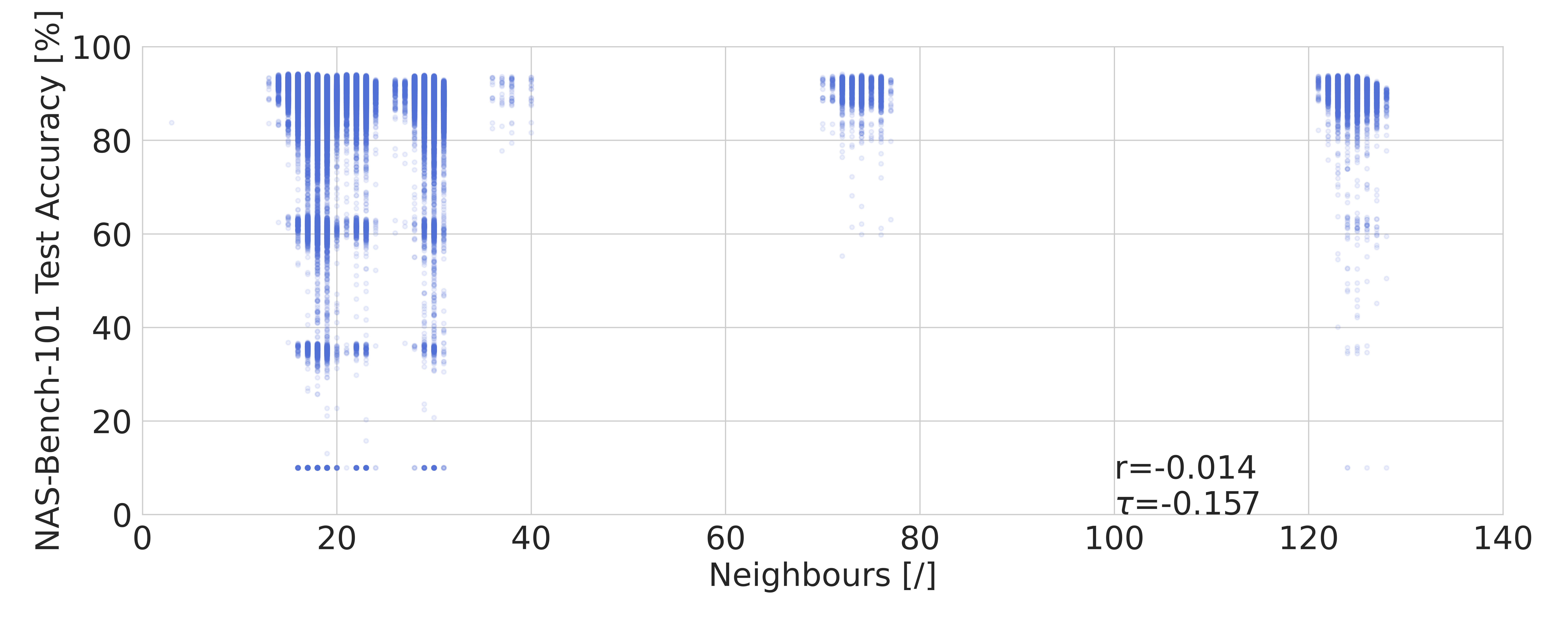}
                \caption{The test accuracy of a NAS-Bench-101 architecture vs. the number of neighbours of the architecture.}
                \label{fig:nb101-test-accuracy-vs-neighbours}
            \end{figure}

            Table \ref{table:nb101-results:best-after-N} shows the numerical results for the best NAS-Bench-101 test accuracy found by our \ac{RL} agent after a specified number of queries. We selected 50 as the best result for a low query budget, 100 as the cross-over point where local search starts outperforming our algorithm, 150 as the halfway point to 300 queries, and 300 since its the most commonly used value by most state-of-the-art methods.

			\begin{sidewaystable}
				\begin{tabular}{l | l | l | r | l | l | r | l | l | r | l | l | r}
					\toprule
								&	\multicolumn{3}{|c|}{50 Queries}	& \multicolumn{3}{|c|}{100 Queries}	& \multicolumn{3}{|c|}{150 Queries}	& \multicolumn{3}{|c}{300 Queries}\\
					Algorithm	&	\#Q	&	$\mu\pm\sigma$	&	95\% CI	&	\#Q	&	$\mu\pm\sigma$	&	95\% CI	&	\#Q	&	$\mu\pm\sigma$	&	95\% CI	&	\#Q	&	$\mu\pm\sigma$	&	95\% CI	\\
					\midrule[\heavyrulewidth]
					Random (NASWOT) \cite{Mellor2021Neural}	& N/A & & & 100 & \makecell{90.38\%\\$\pm$ 5.51\%} & \makecell{[79.55\%,\\101.21\%]} & N/A & & &	N/A	& &		\\
					\hline
					NASWOT \cite{Mellor2021Neural}				& N/A & & & 100 & \makecell{91.77\%\\$\pm$ 0.05\%} & \makecell{[91.67\%,\\91.87\%]} & N/A & & &	N/A	& &		\\
					\hline
					Random (GraphPNAS) \cite{Li2022Graphpnas}	& N/A & & & N/A & & & N/A & & &	300	&	\makecell{93.59\%\\$\pm$ 0.42\%}	&	\makecell{[92.63\%,\\94.54\%]}	\\
					\hline
					NAO	\cite{Luo2020Semi}						& N/A & & & N/A & & & N/A & & &	300	&	\makecell{93.69\%\\$\pm$ 0.06\%}	&	\makecell{[93.57\%,\\93.81\%]}	\\
					\hline
					SemiNAS	\cite{Luo2020Semi}					& N/A & & & N/A & & & N/A & & &	300	&	\makecell{93.89\%\\$\pm$ 0.06\%}	&	\makecell{[93.77\%,\\94.01\%]}	\\
					\hline
					GraphPNAS \cite{Li2022Graphpnas}			& N/A & & & N/A & & & N/A & & &	300	&	\makecell{\bfseries 94.19\%\\\bfseries $\pm$ 0.07\%}	&	\makecell{[94.03\%,\\94.36\%]}	\\
					\midrule[\heavyrulewidth]
					Random Search (Ours)& 50	& \makecell{93.48\%\\$\pm$ 0.15\%}	& \makecell{[93.47\%,\\93.50\%]}	& 100	& \makecell{93.56\%\\$\pm$ 0.14\%}	& \makecell{[93.55\%,\\93.58\%]}	& 150	& \makecell{93.61\%\\$\pm$ 0.15\%}	& \makecell{[93.60\%,\\93.63\%]}	& 300	& \makecell{93.69\%\\$\pm$ 0.16\%}	& \makecell{[93.68\%,\\93.71\%]}	\\
					\hline
					Random Walk (Ours)& 50		& \makecell{93.62\%\\$\pm$ 0.20\%}	& \makecell{[93.60\%,\\93.64\%]}	& 100	& \makecell{93.73\%\\$\pm$ 0.18\%}	& \makecell{[93.71\%,\\93.74\%]}	& 150	& \makecell{93.79\%\\$\pm$ 0.17\%}	& \makecell{[93.78\%,\\93.80\%]}	& 300	& \makecell{93.88\%\\$\pm$ 0.14\%}	& \makecell{[93.87\%,\\93.89\%]}	\\
					\hline
					Local Search (Ours)& 50		& \makecell{93.61\%\\$\pm$ 0.55\%}	& \makecell{[93.57\%,\\93.66\%]}	& 100	& \makecell{\bfseries 93.85\%\\\bfseries $\pm$ 0.33\%}	& \makecell{[93.81\%,\\93.87\%]}	& 150	& \makecell{\bfseries 93.95\%\\\bfseries $\pm$ 0.26\%}	& \makecell{[93.93\%,\\93.97\%]}	& 300	& \makecell{94.05\%\\$\pm$ 0.17\%}	& \makecell{[94.04\%,\\94.07\%]}	\\
					\hline
					Ape-X (Ours)& 50	& \makecell{\bfseries 93.73\%\\ \bfseries $\pm$ 0.16\%}	& \makecell{[93.71\%,\\93.74\%]}	& 100	& \makecell{93.82\%\\$\pm$ 0.15\%}	& \makecell{[93.80\%,\\93.83\%]}	& 150	& \makecell{93.86\%\\$\pm$ 0.14\%}	& \makecell{[93.85\%,\\93.87\%]}	& 300	& \makecell{93.94\%\\$\pm$ 0.13\%}	& \makecell{[93.93\%,\\93.95\%]}	\\
					\botrule
                \end{tabular}
                \caption{Numerical test accuracy values on NAS-Bench-101 (mean $\pm$ std.) and 95\% confidence intervals at various query budgets. Since we may not have data for a specific query number , we use the data of the last preceding improvement.}
                \label{table:nb101-results:best-after-N}
            \end{sidewaystable}

    \subsection{NAS-Bench-301}
        As mentioned at the start of section \ref{sec:experiments}, we will also evaluate our agent on the NAS-Bench-301 setting \cite{Siems2021Nasbench}, due to its increased complexity compared to NAS-Bench-101.

        \subsubsection{Configuration}
            For NAS-Bench-301, our hyperparameter setup is similar to that of NAS-Bench-101. We train our agents for longer on NAS-Bench-301, for up to $15 \times 10^{6}$ timesteps. We sampled a new set of initial states for evaluation (Since NAS-Bench-301 uses a different search space) with the same size as the NAS-Bench-101 set of initial states. We set reward shaping exponent to $32$, rather than $6$ as was the case with NAS-Bench-101. The NAS-Bench-301 agent also uses a 512-dimensional latent space, compared to 256 dimensions for NAS-Bench-101, due to the increased size of the observations, and the fact that it needs to encode two architectures (Normal and Reduction Cell) in one latent space. We use version 1.0 of NAS-Bench-301, with the included ensemble of XGB estimators, without using noisy predictions (i.e., taking the mean of all predictors in the ensemble), following the recommendation from \cite{White_2021_Exploring} to reduce noise in architecture evaluation pipelines (Denoised setting). We made some minor alterations to the original NAS-Bench-301 code to allow for batched inference of the estimator to speed up the training procedure. All experiments were carried out 1 NVIDIA Tesla V100 GPU. The vectorization of our environment was disabled to reduce the overall time taken for a single training iteration. Since the NAS-Bench-301 XGB ensemble is only trained to predict validation accuracy, we both train and evaluate our agents using the validation accuracy.

        \subsubsection{Results}
            We will be evaluating our agent using the same analysis and metrics that we did for NAS-Bench-101, and the evaluation procedure also remains unchanged.
            \begin{figure}
                \centering
                \includegraphics[width = \textwidth, clip]{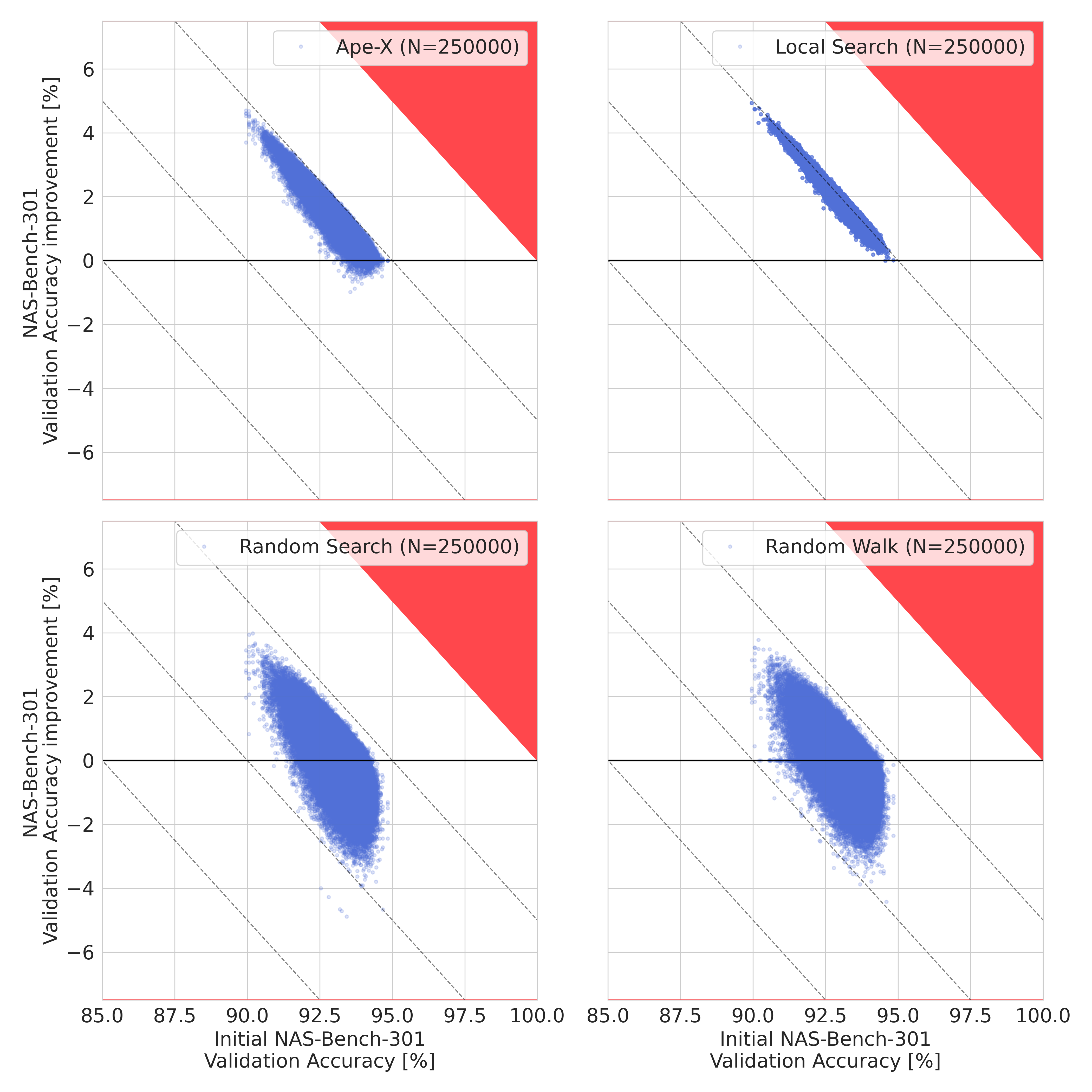}
                \caption{The initial accuracy of the agent vs the improvement in accuracy.}
                \label{fig:nb301-results:initial-vs-improvement}
            \end{figure}

            One thing that is immediately noticeable in the improvement histograms for NAS-Bench-301, is the quasi-normal shaped improvement distribution, for all algorithms considered. This differs considerably from NAS-Bench-101, where the improvement distributions, especially for random search, were much more uniformly distributed.
            To verify this hypothesis, we conducted a normality test with a p-value of $5 \times 10^{-3}$. For all algorithms, the null-hypothesis that the data is normally distributed was rejected, with the test reporting p-values less than $1 \times 10^{-6}$ for all algorithms. We used scipy's ``scipy.stats.normaltest" \cite{Virtanen2020SciPy} test, which combines skew and kurtosis to test the normality of a sample.
            Following these normality tests, we also conducted a test to see if the data might follow a t-distribution. We used ``scipy's scipy.stats.goodness\_of\_fit" \cite{Virtanen2020SciPy} with 1000 Monte-Carlo samples. Once again, we use a p-value of $5 \times 10^{-3}$ as our threshold for rejecting the null hypothesis that our data follows a t-distribution. For all algorithms, we rejected the null hypothesis that the data follows a t-distribution, with each test returning a p-value of $9.99 \times 10^{-4}$ for Ape-X, Local Search and Random Walks, and a p-value of $1.998 \times 10^{-3}$ for Random Search.

            \begin{figure}
                \centering
                \includegraphics[width = \textwidth, clip]{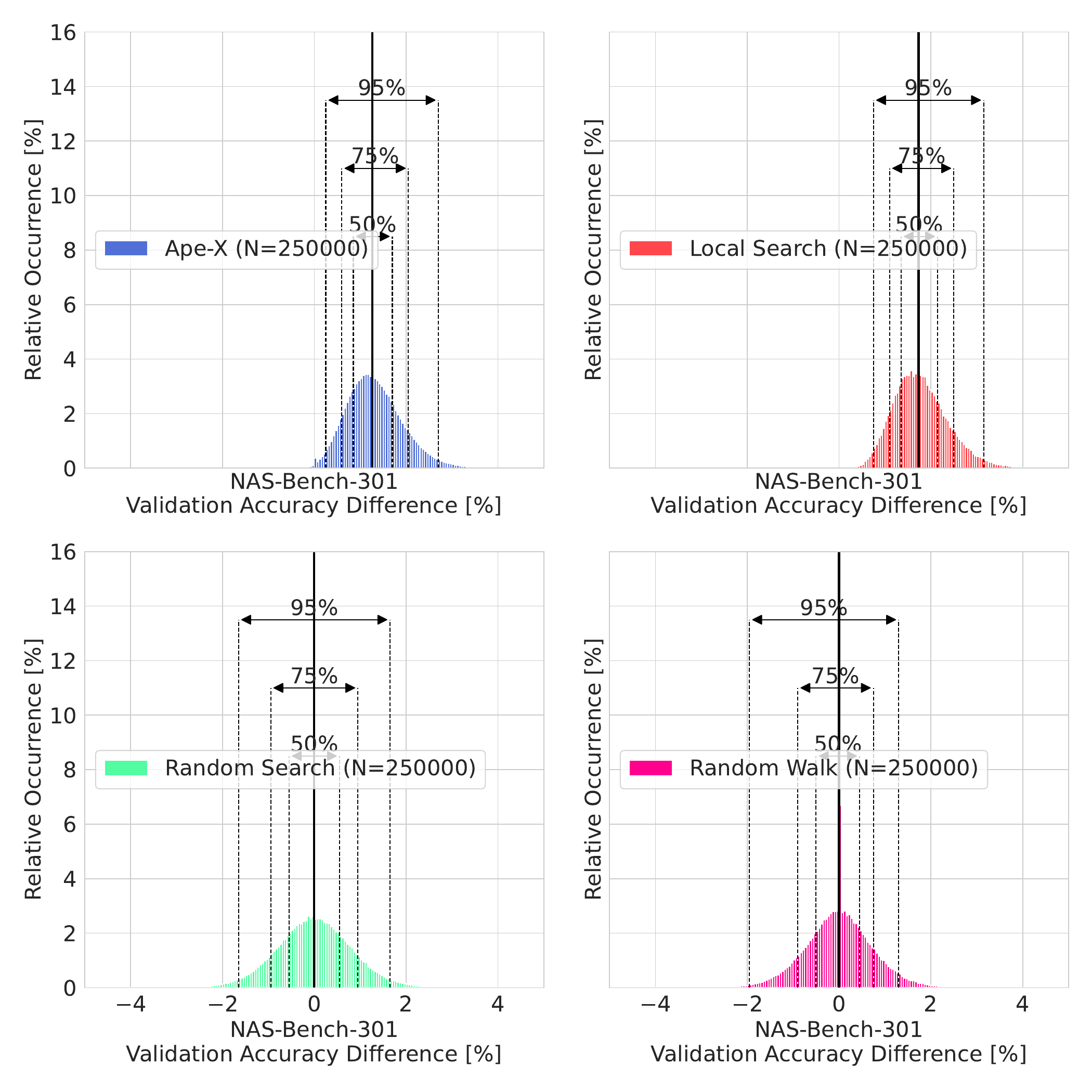}
                \caption{A histogram showing the distribution of the improvement in accuracy over an entire episode}
                \label{fig:nb301-results:improvement-histogram}
            \end{figure}

\begin{figure}
                \centering
                \includegraphics[width = \textwidth, clip]{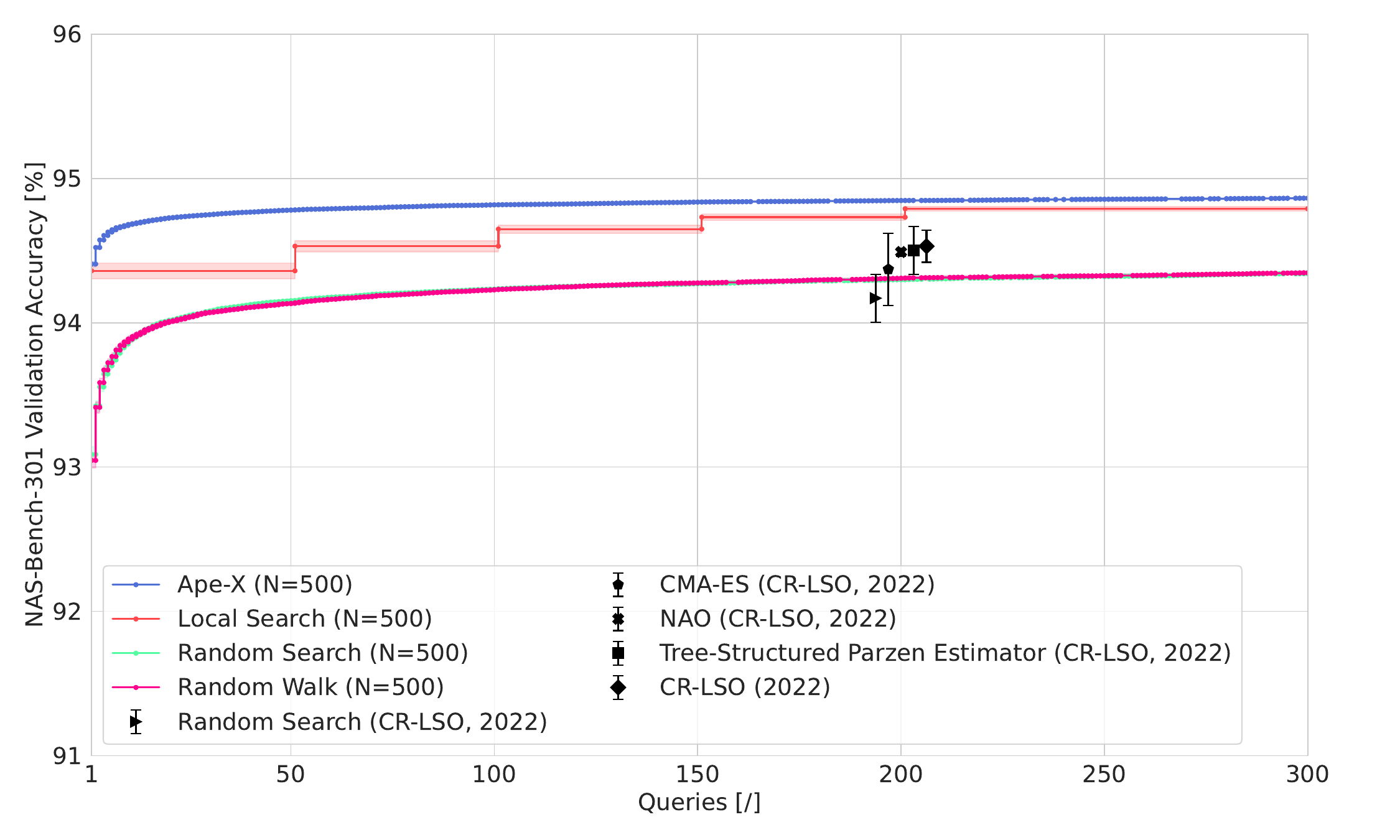}
                \caption{The accuracy of the best architecture found so far plotted against the number of queries. We evaluated each algorithm for up to 300 queries.}
                \label{fig:nb301-results:best-after-N}
            \end{figure}

             The confidence intervals displayed in figure \ref{fig:nb301-results:best-after-N} were computed in the same way as in our NAS-Bench-101 experiments. Figure \ref{fig:nb301-results:best-after-N} shows us a slightly different picture from figure \ref{fig:nb101-results:best-after-N} in the NAS-Bench-101 setting. With a 300-query budget, our \ac{RL} agent actually outperforms local search. Another interesting difference is the near-identical performance of random search and random walks. While in the NAS-Bench-101 setting, these two showed a significant performance difference, in the NAS-Bench-301 case, they both achieve nearly identical performance.

             Since some publications use larger query budgets for NAS-Bench-301 (Ranging from 150 to 10000), we also evaluate our agent with a budget of 1000 queries. When we extend the query budget up to 1000 queries, it becomes clear that, given a sufficiently large computational budget, local search remains a strong search algorithm, surpassing our \ac{RL} algorithm around the 400 query mark. We also see that, just as in the 300-query case, performance for random search and random walks remains nearly identical, suggesting that the difference in performance in the NAS-Bench-101 case can likely be attributed to a difference between the benchmarks.

            \begin{figure}
                \centering
                \includegraphics[width = \textwidth, clip]{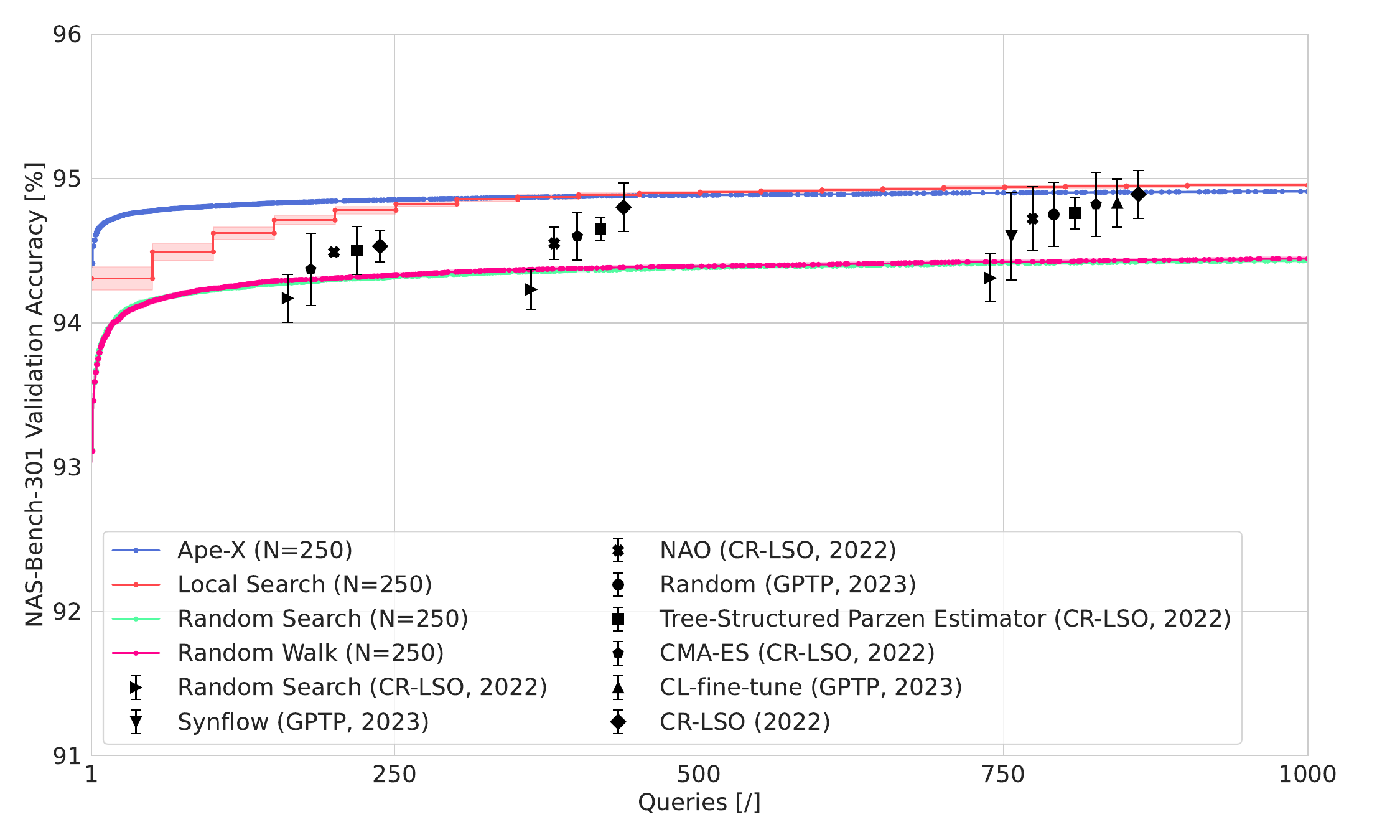}
                \caption{The accuracy of the best architecture found so far plotted against the number of queries. We evaluated each algorithm for up to 1000 queries.}
                \label{fig:nb301-results:best-after-N-1000}
            \end{figure}

            We also provide numerical data on the best mean validation accuracy obtained after a specific number of queries in table \ref{table:nb301-results:best-after-N}. The shown query budgets were chosen  to coincide with the baselines shown in figure \ref{fig:nb301-results:best-after-N}.

			\begin{sidewaystable}
				\begin{tabular}{l | l | l | r | l | l | r | l | l | r | l | l | r}
					\toprule
								&	\multicolumn{3}{|c|}{150 Queries}	& \multicolumn{3}{|c|}{200 Queries}	& \multicolumn{3}{|c|}{300 Queries}	& \multicolumn{3}{|c}{400 Queries}\\
					Algorithm	&	\#Q	&	$\mu\pm\sigma$	&	95\% CI	&	\#Q	&	$\mu\pm\sigma$	&	95\% CI	&	\#Q	&	$\mu\pm\sigma$	&	95\% CI	&	\#Q	&	$\mu\pm\sigma$	&	95\% CI	\\
					\midrule[\heavyrulewidth]
					Random Search (GPTP) \cite{Han2023General}	& N/A & & & N/A & & & N/A & & & N/A & &\\
					\hline
					Synflow (GPTP) \cite{Han2023General}		& N/A & & & N/A & & & N/A & & & N/A & &\\
					\hline
					CL-fine-tune (GPTP) \cite{Han2023General}	& N/A & & & N/A & & & N/A & & & N/A & &\\
					\hline
					Random Search (CR-LSO) \cite{Rao2022CRLSO}	& N/A & & & 200 & \makecell{94.17\%\\$\pm$ 0.06\%} & \makecell{[94.00\%,\\94.34\%]} & N/A & & & 400 & \makecell{94.23\%\\$\pm$ 0.05\%} & \makecell{[94.09\%,\\94.37\%]}\\
					\hline
					TPE (CR-LSO) \cite{Rao2022CRLSO}			& N/A & & & 200 & \makecell{94.50\%\\$\pm$ 0.06\%} & \makecell{[94.33\%,\\94.67\%]} & N/A & & & 400 & \makecell{94.65\%\\$\pm$ 0.03\%} & \makecell{[94.57\%,\\94.73\%]}\\
					\hline
					CMA-ES (CR-LSO) \cite{Rao2022CRLSO}		& N/A & & & 200 & \makecell{94.37\%\\$\pm$ 0.09\%} & \makecell{[94.12\%,\\94.62\%]} & N/A & & & 400 & \makecell{94.60\%\\$\pm$ 0.06\%} & \makecell{[94.43\%,\\94.77\%]}\\
					\hline
					NAO (CR-LSO) \cite{Rao2022CRLSO}			& N/A & & & 200 & \makecell{94.49\%\\$\pm$ 0.01\%} & \makecell{[94.46\%,\\94.52\%]} & N/A & & & 400 & \makecell{94.55\%\\$\pm$ 0.04\%} & \makecell{[94.44\%,\\94.66\%]}\\
					\hline
					CR-LSO \cite{Rao2022CRLSO}					& N/A & & & 200 & \makecell{94.53\%\\$\pm$ 0.04\%} & \makecell{[94.32\%,\\94.64\%]} & N/A & & & 400 & \makecell{94.80\%\\$\pm$ 0.06\%} & \makecell{[94.63\%,\\94.97\%]}\\
					\midrule[\heavyrulewidth]
					Random Search (Ours)	& 149	& \makecell{94.26\%\\$\pm$ 0.12\%}	& \makecell{[94.25\%,\\94.28\%]}	& 198	& \makecell{94.30\%\\$\pm$ 0.11\%}	& \makecell{[94.29\%,\\94.31\%]}	& 297	& \makecell{94.34\%\\$\pm$ 0.09\%}	& \makecell{[94.33\%,\\94.35\%]}	& 400	& \makecell{94.36\%\\$\pm$ 0.09\%}	& \makecell{[94.35\%,\\94.37\%]}\\
					\hline
					Random Walk (Ours)		& 150	& \makecell{94.28\%\\$\pm$ 0.12\%}	& \makecell{[94.26\%,\\94.29\%]}	& 198	& \makecell{94.31\%\\$\pm$ 0.11\%}	& \makecell{[94.29\%,\\94.32\%]}	& 300	& \makecell{94.35\%\\$\pm$ 0.11\%}	& \makecell{[94.33\%,\\94.36\%]}	& 400	& \makecell{94.37\%\\$\pm$ 0.11\%}	& \makecell{[94.35\%,\\94.38\%]}\\
					\hline
					Local Search (Ours)		& 150	& \makecell{94.71\%\\$\pm$ 0.26\%}	& \makecell{[94.68\%,\\94.74\%]}	& 200	& \makecell{94.78\%\\$\pm$ 0.19\%}	& \makecell{[94.75\%,\\94.80\%]}	& 300	& \makecell{94.85\%\\$\pm$ 0.13\%}	& \makecell{[94.84\%,\\94.87\%]}	& 400	& \makecell{\bfseries 94.89\%\\\bfseries$\pm$ 0.11\%}	& \makecell{[94.87\%,\\94.90\%]}\\
					\hline
					Ape-X (Ours)			& 150	& \makecell{\bfseries 94.83\%\\\bfseries$\pm$ 0.05\%}	& \makecell{[94.83\%,\\94.84\%]}	& 199	& \makecell{\bfseries 94.85\%\\\bfseries$\pm$ 0.05\%}	& \makecell{[94.84\%,\\94.85\%]}	& 300	& \makecell{\bfseries 94.86\%\\\bfseries$\pm$ 0.05\%}	& \makecell{[94.86\%,\\94.87\%]}	& 398	& \makecell{94.88\%\\$\pm$ 0.04\%}	& \makecell{[94.87\%,\\94.88\%]}\\
					
					\toprule
								& \multicolumn{3}{|c}{800 Queries} & \multicolumn{3}{|c}{1000 Queries} & & & & & &\\
					Algorithm	&	\#Q	&	$\mu\pm\sigma$	&	95\% CI	&	\#Q	&	$\mu\pm\sigma$	&	95\% CI	& & & & & & \\
					\midrule[\heavyrulewidth]
					Random Search (GPTP) \cite{Han2023General}	& 800	& \makecell{94.75\%\\$\pm$ 0.08\%}	& \makecell{[94.53\%,\\94.97\%]}	& N/A & & & & & & & &\\
					\hline
					Synflow (GPTP) \cite{Han2023General}		& 800	& \makecell{94.60\%\\$\pm$ 0.11\%}	& \makecell{[94.29\%,\\94.91\%]}	& N/A & & & & & & & &\\
					\hline
					CL-fine-tune (GPTP) \cite{Han2023General}	& 800	& \makecell{94.83\%\\$\pm$ 0.06\%}	& \makecell{[94.66\%,\\95.00\%]}	& N/A & & & & & & & &\\
					\hline
					Random Search (CR-LSO) \cite{Rao2022CRLSO}	& 800	& \makecell{94.31\%\\$\pm$ 0.06\%}	& \makecell{[94.14\%,\\94.48\%]}	& N/A & & & & & & & &\\
					\hline
					TPE (CR-LSO) \cite{Rao2022CRLSO}			& 800	& \makecell{94.76\%\\$\pm$ 0.04\%}	& \makecell{[94.65\%,\\94.87\%]}	& N/A & & & & & & & &\\
					\hline
					CMA-ES (CR-LSO) \cite{Rao2022CRLSO}		& 800	& \makecell{94.82\%\\$\pm$ 0.08\%}	& \makecell{[94.60\%,\\95.04\%]}	& N/A & & & & & & & &\\
					\hline
					NAO (CR-LSO) \cite{Rao2022CRLSO}			& 800	& \makecell{94.72\%\\$\pm$ 0.08\%}	& \makecell{[94.50\%,\\94.94\%]}	& N/A & & & & & & & &\\
					\hline
					CR-LSO \cite{Rao2022CRLSO}					& 800	& \makecell{94.89\%\\$\pm$ 0.06\%}	& \makecell{[94.72\%,\\95.06\%]}	& N/A & & & & & & & &\\
					\midrule[\heavyrulewidth]
					Random Search (Ours)	& 798	& \makecell{94.41\%\\$\pm$ 0.08\%}	& \makecell{[94.40\%,\\94.42\%]}	& 1000	& \makecell{94.43\%\\$\pm$ 0.08\%}	& \makecell{[94.42\%,\\94.44\%]} & & & & & &\\
					\hline
					Random Walk (Ours)		& 797	& \makecell{94.42\%\\$\pm$ 0.10\%}	& \makecell{[94.41\%,\\94.43\%]}	& 1000	& \makecell{94.44\%\\$\pm$ 0.09\%}	& \makecell{[94.43\%,\\94.45\%]} & & & & & &\\
					\hline
					Local Search (Ours)		& 800	& \makecell{\bfseries 94.94\%\\\bfseries $\pm$ 0.09\%}	& \makecell{[94.93\%,\\94.96\%]}	& 1000	& \makecell{\bfseries 94.95\%\\\bfseries$\pm$ 0.08\%}	& \makecell{[94.94\%,\\94.96\%]} & & & & & &\\
					\hline
					Ape-X (Ours)			& 800	& \makecell{94.90\%\\$\pm$ 0.04\%}	& \makecell{[94.90\%,\\94.91\%]}	& 1000	& \makecell{94.91\%\\$\pm$ 0.04\%}	& \makecell{[94.90\%,\\94.91\%]} & & & & & &\\
					\botrule
                \end{tabular}
                \caption{Numerical validation accuracy values on NAS-Bench-301 (mean $\pm$ std.) and 95\% confidence intervals at various query budgets. Since we may not have data for a specific query number, we use the data of the last preceding improvement.}
                \label{table:nb301-results:best-after-N}
            \end{sidewaystable}
			
			Finally, we consider the time it takes to train and evaluate our \ac{RL} agent. The mean training time for NAS-Bench-301 was $486.16$h ($\sigma=40.12$h), with a minimum of $446.75$h, and a maximum of $553.48$h. During evaluation, using only CPUs (no GPUs), completing an episode took an average of $16.30$s ($\sigma=9.52$ s), with a minimum of $0.00$s and a maximum of $69.83$s.
            
			\begin{figure}
                \centering
                \includegraphics[width = \textwidth, clip]{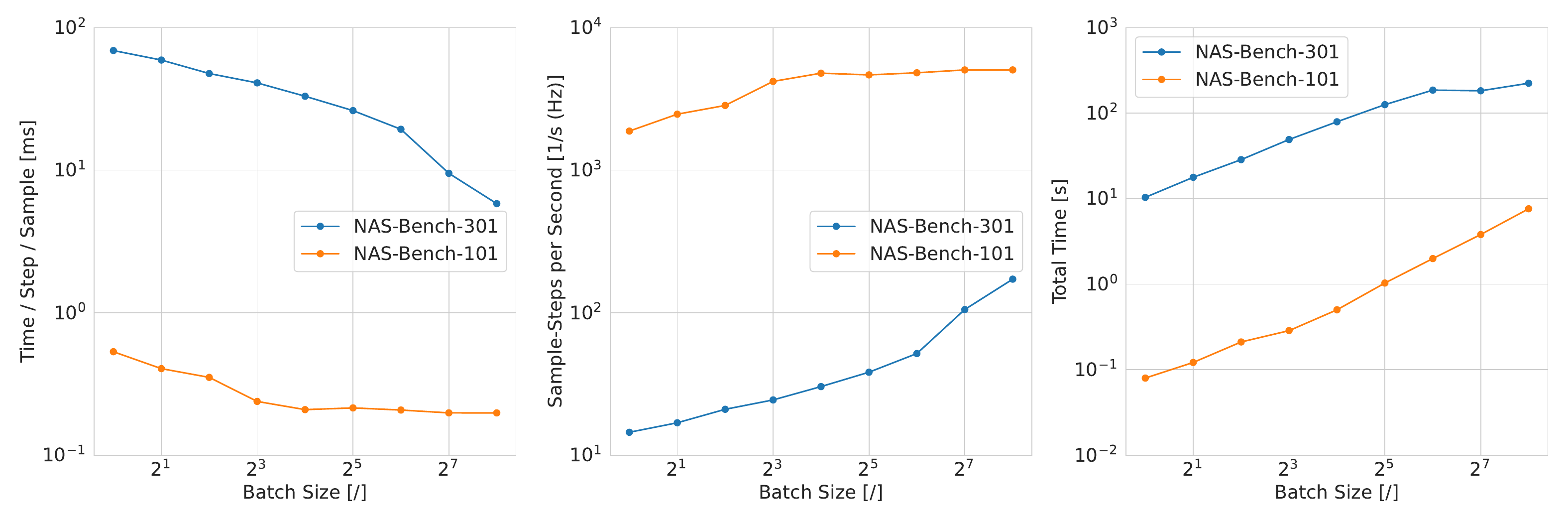}
                \caption{A comparison of the throughput of our environment for NAS-Bench-101 and NAS-Bench-301, with various degrees of vectorization. Note the logarithmic axes.}
                \label{fig:env-performance}
            \end{figure}            
            
            We note an important difference in the training and evaluation time between the NAS-Bench-101 and NAS-Bench-301 setting. The average training time for NAS-Bench-101 is around 4 days, while that for NAS-Bench-301 is around 21 days ($5.25\times$), despite NAS-Bench-301 only requiring $1.5\times$ as much training experience, resulting in the conclusion that in the NAS-Bench-301 setting, we are on average $3.5\times$ slower to train. When looking at the evaluation times, we see an even starker contrast of 0.5s/Episode for NAS-Bench-101, and 16.3s/Episode for NAS-Bench-301 ($32\times$). We attribute this difference in training time primarly to the time it takes to evaluate both reward functions. Since evaluating a reward function is a fairly common occurrence in \ac{RL} (It occurs at least once per timestep), a slow reward function will inevitably slow down the whole training process. Because NAS-Bench-301 requires the evaluation of 10 different gradient boosted trees, its reward function is significantly slower than that of NAS-Bench-101, which consists of simply a table look-up. We can also see this in figure \ref{fig:env-performance}, which compares the performance of our environment when using the NAS-Bench-101 objective versus the NAS-Bench-301 objective. To generate these numbers, we stripped our environment down to its bare essentials, so we can focus solely on the effect that these objectives have on the throughput. The environment no longer uses the incremental formulation we've used so far, and no longer need to do things like generate neighbours. We simply feed the environment an architecture, as our action, and it returns the reward. We can clearly see that, when using the NAS-Bench-101 objective, our environment tends to be around two orders of magnitude quicker than when using NAS-Bench-301, regardless of the degree of vectorization used.

\section{Ablation Studies}

    We also perform several ablation studies to study the effect of certain parameter choices on our overall system. We use the NAS-Bench-101 setting for this, since it is the quickest to train, allowing us to conduct more experiments. We compare different parameter choices by looking at the mean validation accuracy of the architectures produced throughout the training procedure.

	\subsection{Gamma}
\begin{figure}
			\centering
            \includegraphics[width = \textwidth, clip]{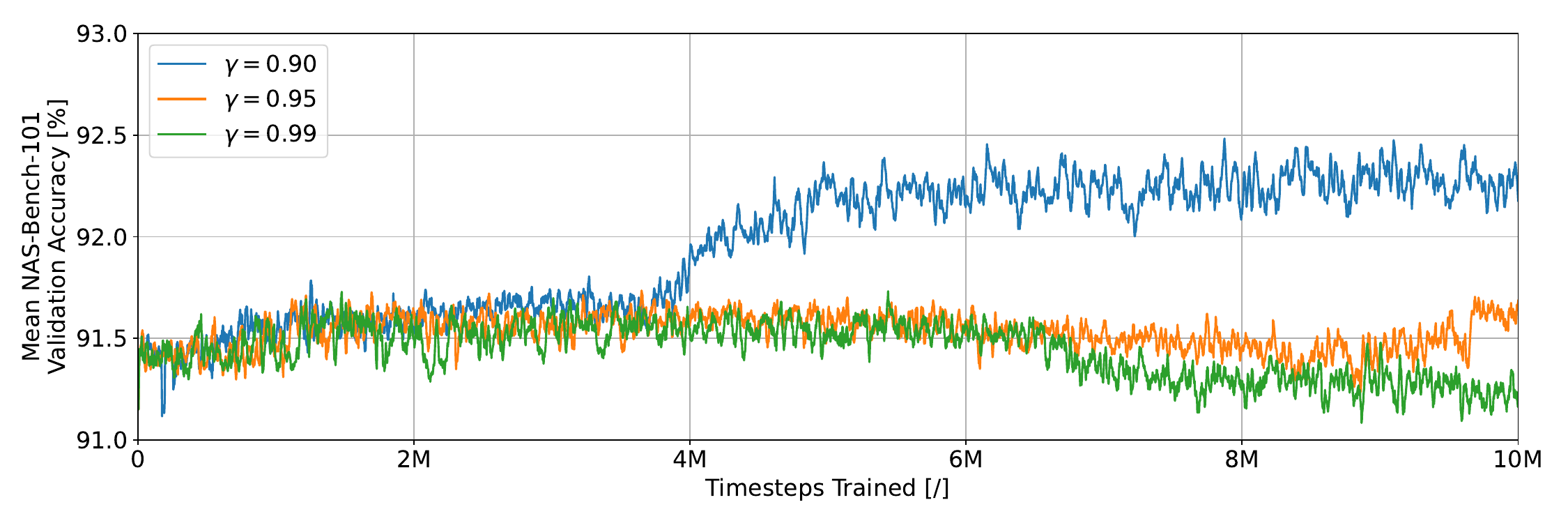}
            \caption{Mean validation accuracy over time for $\gamma = 0.9, 0.95 \textrm{ and } 0.99$ on NAS-Bench-101. A simple moving average filter with N=10 was used to smooth out the results.}
            \label{fig:ablation:gamma}
        \end{figure}

       First, we consider the $\gamma$ parameter that is used in \ac{RL} to determine how heavily future rewards should be discounted compared to current rewards. Lower values for $\gamma$ discount future rewards more, and thus focus more on immediate rewards, and less on future rewards. We consider three values for $\gamma: 0.9, 0.95 \textrm{ and } 0.99$. In figure \ref{fig:ablation:gamma} we show the mean accuracy of the architectures generated as a function of the number of timesteps the agent has been trained on. The figure shows that, during the training procedure, only the agent trained with $\gamma = 0.9$ is able to make a noticable improvement. The agents with $\gamma = 0.95 \textrm{ and } 0.99$ both fail to significantly improve the accuracy of the generated architectures, and the agent with $\gamma = 0.99$ even ends up with slightly worse performance. During training, we noted that, on average, every architecture has around 25 neighbours, this means that, N steps into the future, an agent can be at any of $25^{N}$ different states. Since the number of possible states that can be reached in relatively few steps grows very quickly, making an accurate determination of the future discounted reward becomes very difficult, thus hindering convergence when larger values of $\gamma$ are used.

    \subsection{Number of Neighbours}
        Our environment limits the number of neighbours that our agent can choose from. The agent is presented with up $N$ neighbours, or less if the total number of neighbours is less than $N$. We consider three values for this parameter: $25, 50 \textrm{ and } 100$.
        From this, we can see that agents observing 25 or 50 neighbours both converge to roughly the same performance, while observing up to 100 neighbours leads to slightly worse performance. We hypothesize that this is likely a function of model capacity. The same model was used in all three experiments, but the size of the observations, and the number of architectures that needs to be considered quadrupled between $N=25$ and $N=100$. These results also show that, in a search space such as NAS-Bench-101 which is fairly densely connected, raising the total number of neighbours beyond a certain threshold doesn't help agent performance.

        \begin{figure}
            \centering
            \includegraphics[width = \textwidth, clip]{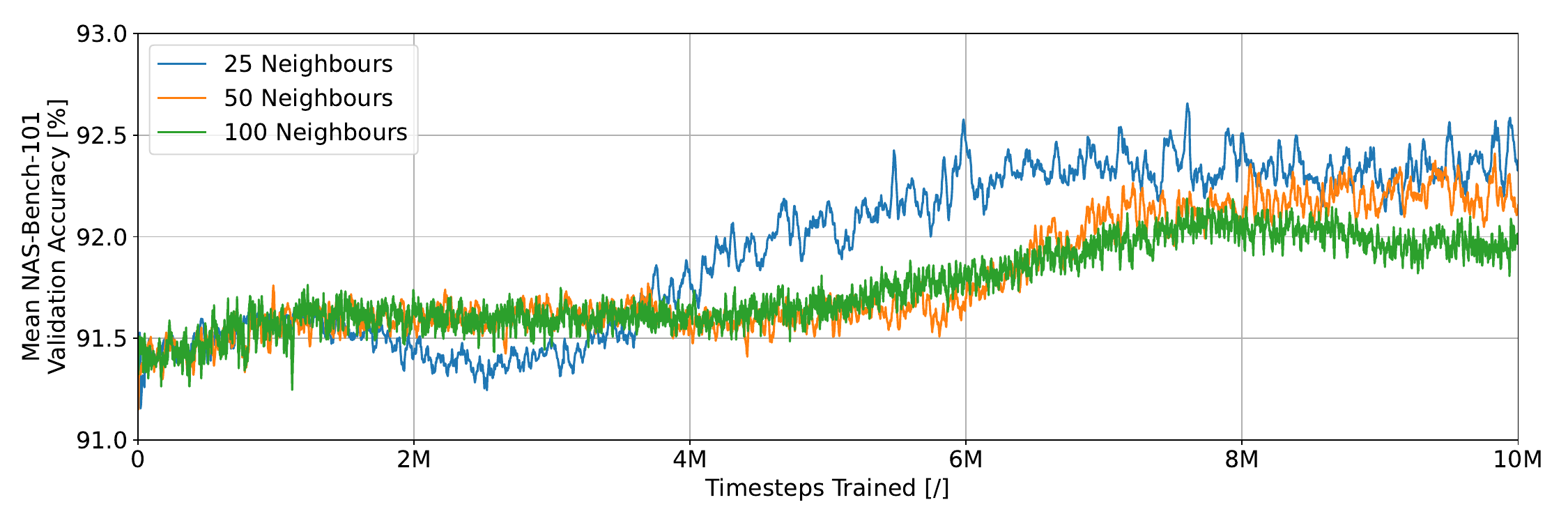}
            \caption{Mean validation accuracy over time for $N = 25, 50 \textrm{ and } 100$ on NAS-Bench-101. A simple moving average filter with N=10 was used to smooth out the results.}
            \label{fig:ablation:neighbours}
        \end{figure}

    \subsection{Reward Shaping}
\begin{figure}
            \centering
            \includegraphics[width = \textwidth, clip]{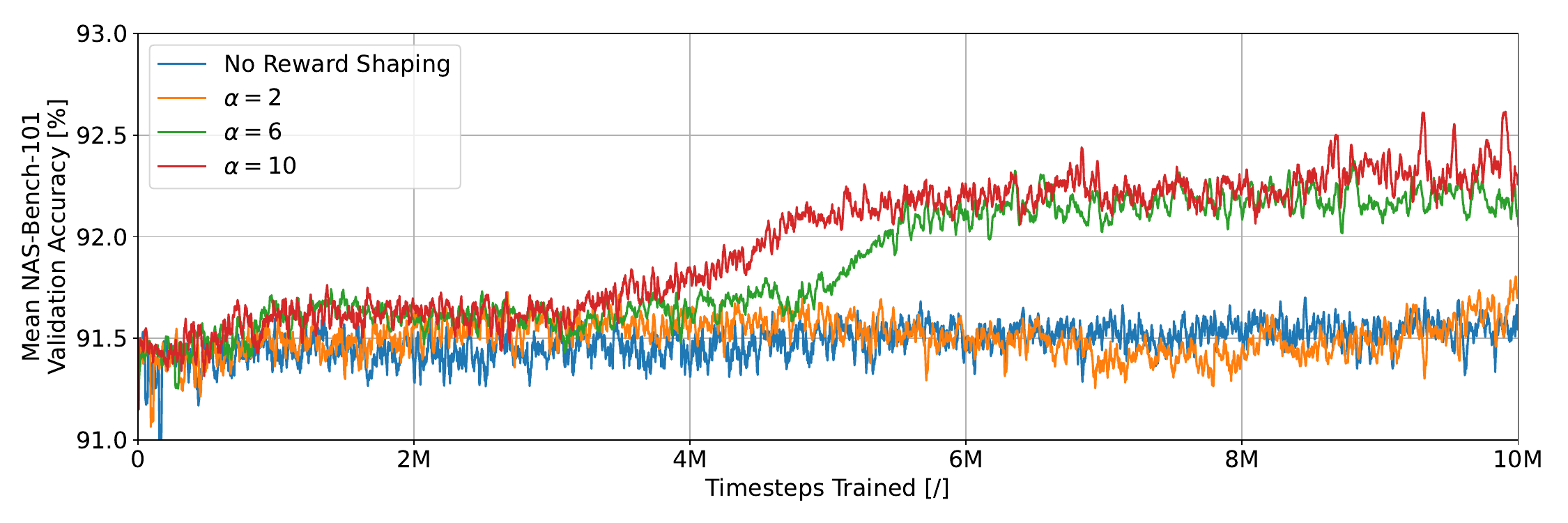}
            \caption{Mean validation accuracy over time for no reward shaping, and $\alpha = 2, 6 \textrm{ and } 10$ on NAS-Bench-101. A simple moving average filter with N=10 was used to smooth out the results.}
            \label{fig:ablation:reward_shaping}
        \end{figure}

        In the NAS-Bench-101 setting, the majority of architectures have an accuracy around 90\%, with only a small minority having accuracies far below 90\%.
        Because of this, we introduced a reward shaping scheme in section \ref{ssec:reward-shaping}. In this section, we will consider the effect that different values of the shaping parameter $\alpha$ have on the performance of our agent. We consider four settings for this reward shaping mechanism: No reward shaping (No exponential functions are applied), and three values for the reward shaping coefficient: $2, 6 \textrm{ and } 10$. The results of this experiment are shown in figure \ref{fig:ablation:reward_shaping}. In this figure, we see that the experiments without reward shaping, or with only slight reward shaping ($\alpha=2$) achieve worse performance than experiments with stronger reward shaping ($\alpha=6, 10$). From this, we conclude that, in order to achieve good performance, a sufficient level of reward shaping is necessary. We also note that the experiment with $\alpha=10$ is slightly quicker to experience a rise in accuracy, however, we do not believe this to be significant, since in our ablations for $\gamma$, we used $\alpha=6$, and saw results similar to $\alpha=10$ when $\gamma=0.9$.
        
\section{Conclusions}
	In this publication, we outline a new, incremental framing of the \ac{NAS} problem. We also introduce a method of tackling this problem using a \ac{RL}-based \ac{NAS} agent. We show that our \ac{RL} agent is competitive with state-of-the-art \ac{NAS} algorithms and baselines that are known to have strong performance. Our \ac{NAS} agent outperforms all other benchmarks considered when query budgets are low, but starts to be overtaken by other algorithms as query budgets increase. We also perform an ablation study and conclude that our agent has a high sensitivity to some hyperparameters, while being rather insensitive to others.

	\subsection{Training Times}
		When comparing training times between the NAS-Bench-101 and NAS-Bench-301 settings, we note a significant difference between both ($\approx5\times$) in terms of training time, even after accounting for the additional training required for NAS-Bench-301 ($\approx3.5\times$). We attribute this difference primarily to the difference in time it takes to evaluate each reward function, with NAS-Bench-301 being significantly slower, due to the need to evaluate 10 gradient boosted trees. This presents a significant barrier to the adoption of our \ac{RL} based method, but it can be overcome by improving the sample-efficiency of the \ac{RL} agent, or using surrogate models that are quicker to evaluate than the ensemble of gradient boosted trees used in this paper.

    \subsection{Random Search}
        In the NAS-Bench-101 setting, we observe a noticable difference between the random search and random walk algorithms. When comparing both algorithms in the NAS-Bench-301 setting, we note that they achieve near identical performance. This indicates that the difference in performance between both algorithms is likely a result of a pecularity of the NAS-Bench-101 benchmark. This does draw into question the validty of comparing the results of algorithms across problem formulations, since, in some cases, formulating the problem in a different way can lead to significantly different performance, even for random algorithms.

    \subsection{Scalability}
        Despite an increase in the size of the search space from $4.23 \times 10^{5}$ to $10^{18}$, the amount of samples required to train our \ac{RL} agent only increased from $1.0 \times 10^{7}$ to $1.5 \times 10^{7}$, while retaining strong performance in both benchmarks. This shows that a \ac{RL}-based approach can scale well as the search space we operate in becomes larger.

	\subsection{Ablation Studies}
		Our ablation studies show mixed results with regards to robustness to hyperparameter changes. Some hyperparameters (like the number of neighbours the agent is presented with) seem to have relatively little effect on the agent's ability to converge, while others (such as the degree of reward shaping and reward discounting) seem to have an almost-binary effect on convergence: Sufficiently high or low values must be selected in order to ensure convergence of the agent.

\section{Future Work}
    \subsection{Re-usability evaluation}
        An important distinction between our transformer-based controller and earlier \ac{RL}-based \ac{NAS} controllers is its re-usability. The traditional \ac{RL}-based \ac{NAS} paradigm involves training an \ac{RL} agent to output a single architecture. This means that, once the training procedure is finished, the optimal architecture is found, but the \ac{RL} agent serves no purpose beyond this point. Accordingly, the time and compute cost for training this agent has gone entirely towards finding a single optimal architecture. Through the re-use our agent offers, this time-cost can be amortized over many searches, since we didn't learn the optimal solution, but rather how to find it.

    \subsection{Domain Generalization}
        In this work, we onyl considered the image classification domain. This unfortunately means that a new agent must be trained every time we wish to tackle a new problem domain. A truly re-usable \ac{NAS} agent should be re-usable on a completely new domain with little to no adaptation work. There are several avenues that could enable this. One of these, is the use of a training-free, domain-independent performance estimator such as Neural Tangent Kernel-based metrics like \ac{LGA} \cite{Mok2022Demystifying}, or the number of linearly separable regions \cite{Mellor2021Neural}. Using metrics like these as performance estimators would create an agent that can theoretically operate in any search space, regardless of the target domain, assuming that the used metrics correlate strongly with actual domain-performance across domains.

\backmatter

\bmhead{Acknowledgments}

\acknowledgement

\begin{appendices}

\end{appendices}

\bibliography{references}

\end{document}